\documentclass[pdflatex,sn-mathphys-num]{sn-jnl}

\usepackage{microtype}
\usepackage{amsmath}
\usepackage{graphicx}
\usepackage{booktabs}
\usepackage{tabularx}
\usepackage{array}
\usepackage{enumitem}
\usepackage{xcolor}
\usepackage{fvextra}
\usepackage{url}
\hypersetup{hidelinks,bookmarksdepth=2,hypertexnames=false}
\usepackage[nameinlink,noabbrev]{cleveref}
\definecolor{promptframe}{gray}{0.72}
\DefineVerbatimEnvironment{PromptText}{Verbatim}{%
  fontsize=\scriptsize,
  breaklines=true,
  breakanywhere=true,
  breaksymbolleft={},
  frame=single,
  framesep=2mm,
  rulecolor=\color{promptframe}
}
\fvset{%
  fontsize=\scriptsize,
  breaklines=true,
  breakanywhere=true,
  breaksymbolleft={},
  frame=single,
  framesep=2mm,
  rulecolor=\color{promptframe}
}

\newcommand{\twma}{\texorpdfstring{\texttt{twma\_\allowbreak v1.2}}{twma_v1.2}}
\newcommand{\ewma}{\texorpdfstring{\texttt{ewma\_\allowbreak v1.2}}{ewma_v1.2}}
\newcommand{\ewmas}{\texorpdfstring{\texttt{ewma\_\allowbreak s\_\allowbreak v1.2}}{ewma_s_v1.2}}
\newcommand{\ewmasv}{\texorpdfstring{\texttt{ewma\_\allowbreak sv\_\allowbreak v1.2}}{ewma_sv_v1.2}}
\newcommand{\ewmasvonefive}{\texorpdfstring{\texttt{ewma\_\allowbreak sv\_\allowbreak v1.5}}{ewma_sv_v1.5}}
\newcommand{\twmavonesix}{\texorpdfstring{\texttt{twma\_\allowbreak v1.6}}{twma_v1.6}}
\newcommand{\ewmasvonesix}{\texorpdfstring{\texttt{ewma\_\allowbreak sv\_\allowbreak v1.6}}{ewma_sv_v1.6}}
\newcommand{\gptfivefour}{\texorpdfstring{\texttt{gpt-5.4}}{gpt-5.4}}
\newcommand{\gptfivefive}{\texorpdfstring{\texttt{gpt-5.5}}{gpt-5.5}}
\newcommand{\gptfivesixsol}{\texorpdfstring{\texttt{gpt-5.6-sol}}{gpt-5.6-sol}}
\newcommand{\promptfile}[2]{%
  \subsubsection{#1}
  \VerbatimInput{#2}
}

\begin{document}

\title{Do Coding Agents Need Executable World Models, Simplification, and
Verification to Solve ARC-AGI-3?}

\author*{\fnm{Sergey} \sur{Rodionov}}\email{sergey@singularitynet.io}
\affil{\orgname{SingularityNET}, \city{Zug}, \country{Switzerland}}

\abstract{
Our previous ARC-AGI-3 agent bundled executable world modeling, prompted
simplification, and exact replay verification, leaving their individual
contributions unclear. An executable world model
is a persistent, agent-authored environment hypothesis embodied in runnable
code. We compare four Codex-based variants: textual; flexible-interface
executable; executable with simplification prompts; and a fixed-interface
variant with simplification and exact replay verification against recorded
observations. The main study evaluates them with \gptfivefour{} and
\gptfivefive{} at \texttt{high} and \texttt{xhigh} reasoning effort on 25
public games; exploratory follow-ups compare textual and verification with
\gptfivesixsol{}.

In the main study, every variant scores higher as model capability and reasoning
effort increase. These gains often exceed variant differences, which are
smaller than anticipated and vary across settings.
Requiring an executable deliverable is not universally beneficial: textual
outperforms flexible-interface executable in both \gptfivefive{} conditions.
The simplification variant scores higher than its executable-only counterpart
in three of four settings; the weakest is the exception. The complete
verification treatment ranks first throughout, sometimes narrowly, but uses
substantially more resources.

With \gptfivesixsol{}, the verification variant completes every public level at
\texttt{xhigh} and \texttt{max} with about 99\% human-relative action
efficiency while using fewer than half the human baseline's total actions. At
\texttt{max}, however, the textual variant completes every level with 41\%
fewer actions than the human baseline. Thus, at \texttt{max}, the three
imposed mechanisms are not required for action-efficient public-set completion;
verification nevertheless scores higher and succeeds at lower effort. Because
\gptfivesixsol{} postdates the games and held-out performance is untested,
results indicate public-set saturation only.
}


\maketitle

\section{Introduction}
\label{sec:introduction}

General-purpose agents in interactive environments must do more than produce a
correct output from a fixed input. They must decide what evidence to acquire,
infer the environment's dynamics and objective from incomplete observations,
and act while uncertainty remains. ARC-AGI-3 makes this challenge explicit. In
each abstract, turn-based game, the agent must infer the action semantics,
dynamics, and objective from interaction while completing the levels
efficiently. Relative Human Action Efficiency (RHAE) measures level completion
and action efficiency relative to a human first-contact baseline
\cite{arcprize2026arcagi3}. Each action can therefore both advance the game and
test a hypothesis, but every such experiment also counts toward the score.

World models are a natural response to this problem. A model-based agent can
use an environment hypothesis to anticipate the consequences of candidate
actions before taking them in the environment
\cite{sutton1991dyna,ha2018worldmodels}. For a coding agent, one possible
representation of this hypothesis is an executable program. In this article,
an \emph{executable world model} is a persistent, agent-authored hypothesis
about the current task environment embodied in runnable code. The qualifier
\emph{persistent} means that the
program remains in the agent's workspace and is maintained across turns within
a game. Here, \emph{world} refers only to the bounded game environment.

Such a program can be inspected, edited, and executed to support reasoning
about the environment without consuming additional environment actions.
Programmatic world-model systems such as WorldCoder and Code World Models show
that language models can synthesize executable dynamics models from
descriptions or interaction data and use them for planning
\cite{tang2024worldcoder,dainese2024codeworldmodels}. Executability is
nevertheless a systems choice, not a free source of accuracy. Maintaining a
simulator consumes reasoning time, memory, and tool calls, while planning with
an inaccurate simulator can misdirect the agent. A capable coding agent may
instead maintain a textual hypothesis and create small, transient programs only
when useful.

Two additional mechanisms may improve the reliability of executable world
models. The first is replay verification. Language models can be effective
proposal mechanisms when their outputs can be checked externally, as in
AlphaCode, FunSearch, and AlphaGeometry
\cite{li2022alphacode,romeraparedes2024funsearch,trinh2024alphageometry}.
For an initially unknown interactive environment, however, a complete verifier
for a proposed world model is generally unavailable. The recorded interaction
history can nevertheless provide a partial test: past actions are replayed in
the model without additional environment interaction, and the resulting
predictions must match the corresponding recorded observations exactly. A
match does not show that the model will be accurate in unobserved situations,
but a mismatch provides concrete evidence for revision.

The second is a simplicity bias. When many hypotheses fit a short trajectory,
Occam's razor favors simpler ones as more likely to generalize beyond the
observed cases. Minimum Description Length gives this intuition a
compression-based formulation \cite{grunwald2004mdl}. We implement this bias as
a scheduled simplification treatment in which the controller asks the agent to
replace special cases with shared rules. This is a practical proxy for an
MDL-style bias, not formal MDL optimization. Both mechanisms also have potential
costs: exact replay may direct effort toward incidental visual details, while
premature simplification may damage a partially correct model.

Recent evidence from ARC-AGI-3 agent traces provides concrete motivation for
externalizing and testing environment hypotheses. An ARC Prize analysis of
160 \gptfivefive{} and Claude Opus 4.7 traces identified cases in which agents
recognized a local action effect but did not integrate it into a correct global
rule, imported an inappropriate abstraction from a familiar game, or completed
an early level without learning a rule that transferred to later levels
\cite{kamradt2026modelanalysis}. That analysis did not compare representation
or verification choices, so it does not establish that a persistent and
testable artifact would solve these problems. It does, however, identify
failures of evidence integration, abstraction choice, and cross-level transfer
that a persistent and testable model is intended to address.

ARC-AGI-3 agents span a spectrum of harness designs. The Duck Harness uses a
lightweight coding loop around a Python REPL \cite{bessis2026duck}, while other
systems externalize more structure: graph-based exploration records observed
states and transitions \cite{rudakov2025graph}; the Agentica team's Arcgentica
harness and DreamTeam distribute exploration, modeling, testing, and planning
across specialized roles
\cite{symbolica2026agentica,symbolica2026arcgentica,sarafian2026workspace}; and
OPINE-World separates acting from program synthesis and checks candidate
object-centric transition models through exact replay of recorded object-state
transitions \cite{courtis2026opineworld}. Because these systems also differ in
foundation model, prompting, memory, orchestration, representation, planning,
verification policy, and evaluation protocol, their end-to-end results do not
isolate the effects of executability, simplification, or verification.

The ARC-AGI-3 authors explicitly designate the 25 public environments as an
easier demonstration set and reserve semi-private and fully private sets for
claims about generalization; they also caution against treating self-reported
community scores as evidence of progress toward AGI
\cite{arcprize2026arcagi3}. We use the public set here as a common experimental
substrate for controlled architectural comparisons, not as a substitute for
sealed evaluation. Within that scope, the literature leaves a basic question
open: when the underlying coding agent is already capable, does requiring a
persistent executable environment hypothesis help, and what additional value
comes from explicit simplification and replay verification?

Our first article combined executable representation, scheduled
simplification, exact replay verification, and model-based plan execution in a
single proof-of-concept agent \cite{rodionov2026executableworldmodels}. Its
results showed that the combined design could work, but left a central
attribution question unanswered: \emph{which parts of the combination are
actually responsible for the performance?} The present article turns that
proof of concept into a nested ablation study that successively adds executable
modeling, simplification, and verification to a textual-model baseline.
All four variants retain shell and Python access, so the comparison concerns
progressively added executable-modeling, simplification, and verification
requirements rather than access to code itself.

\section{Related Work and Conceptual Positioning}
\label{sec:related-work}

\subsection{World models and programmatic environment models}

Model-based agents use predictions of action consequences to plan before acting
in the real environment. Dyna provides an early architecture that integrates
learning from real interaction with planning from simulated experience
\cite{sutton1991dyna}. Modern neural world models generally learn latent or
task-relevant predictive representations. Ha and Schmidhuber learn a compressed
generative model in which a controller can be trained
\cite{ha2018worldmodels}, while MuZero learns the reward, value, and policy
predictions needed for planning without reconstructing the complete observation
\cite{schrittwieser2020muzero}.

The executable models studied here occupy a different point in this design
space. They are external Python artifacts written by a pretrained coding agent
during one game, without gradient updates or a separate model-training phase.
They can be inspected, edited, executed, and tested. Exact replay verification
additionally requires initial-state reconstruction and an observation renderer
so that model predictions can be compared exactly with recorded observations.
A directly relevant line of prior work is programmatic world modeling.
WorldCoder builds a Python environment model from interaction, revises it when
it is inconsistent with collected transitions, and uses it for planning, while
Code World Models generate and repair Python dynamics models using language
models and Monte Carlo tree search
\cite{tang2024worldcoder,dainese2024codeworldmodels}. These systems establish
that programmatic environment models can be constructed, checked against
experience, and used for planning. Our question is whether requiring such a
persistent artifact benefits an otherwise capable coding agent.

\subsection{Program induction, simplification, and verification}

The construction of an executable environment hypothesis is related to program
induction, in which programs are inferred from examples or specifications.
DreamCoder learns program solutions and reusable symbolic abstractions, while
LILO combines language-model-guided synthesis with library learning and
compression \cite{ellis2020dreamcoder,grand2023lilo}. Our setting does not
introduce a corresponding synthesis algorithm: the agent writes unrestricted
Python online from a sequential action--observation history, without searching
within a domain-specific language or optimizing a formal Bayesian or
description-length objective. The scheduled simplification prompts apply a
practical simplicity pressure inspired by MDL and code refactoring
\cite{grunwald2004mdl,kovacic2025refactoring}; they do not perform formal MDL
optimization or cross-task library learning.

Replay verification is related to generate-and-verify systems, in which a
model generates candidates that are checked externally, as in AlphaCode,
FunSearch, and AlphaGeometry
\cite{li2022alphacode,romeraparedes2024funsearch,trinh2024alphageometry}. In our
setting, however, no complete specification of an ARC-AGI-3 environment is
available. Our verifier checks the candidate model only against the finite
recorded interaction history. Passing replay therefore establishes consistency
with observed transitions, not correctness for unobserved states or action
sequences.

\subsection{Coding agents}

LLM-based agents for software engineering extend language models with access to
external resources and tools \cite{liu2026agentsurvey}. All four variants in our
study use the same coding-agent runtime, with filesystem, shell, and Python
access. The treatments differ in whether the agent must maintain its environment
hypothesis as a persistent executable artifact and, if so, how that artifact is
structured, simplified, and tested. Thus, the experiment compares
environment-modeling requirements within a coding-agent system, not code use
with no code use.

\subsection{Model-based reasoning with language models}

Model-based language-reasoning systems provide another comparison. RAP uses a
language model both as a reasoning agent and as a predictive world model within
Monte Carlo tree search \cite{hao2023rap}. LLM+P instead translates a
natural-language description of the initial state and goal into a formal
planning problem, using a supplied domain model that specifies the available
actions and their preconditions and effects, and then invokes a classical planner
\cite{liu2023llmp}. In ARC-AGI-3, the action semantics, transition rules, and
objective are initially unknown and must be inferred through interaction. In
our executable variants, the coding agent instead maintains its environment
hypothesis as a persistent Python artifact that can be inspected, executed, and
tested.

Taken together, these comparisons position executable modeling here as a
representation and orchestration requirement within a coding-agent system.
We evaluate whether that requirement is beneficial, and what additional value
comes from simplification and replay verification, through the nested treatment
contrasts defined in \cref{sec:study-design}.

\section{Study Design}
\label{sec:study-design}

The first article bundled three mechanisms that the present design separates.
First, the agent represented the game's mechanics as executable code, which may
provide a more precise and operational representation than text alone. The
resulting model can be executed, tested,
and used to explore possible action sequences. However, maintaining a simulator
may consume substantial reasoning and interaction time, and a capable coding
agent might instead solve the task using only a textual description of the game,
supplemented by small ad hoc scripts when needed. Second, the controller
repeatedly asked the agent to simplify and
generalize its model. This may provide a useful simplicity bias, but repeated
refactoring can also destabilize a partially correct model. Third, the full
agent was asked to reproduce recorded observations exactly and was given
verifier programs that exposed model--observation mismatches. These mismatches may draw
attention to otherwise overlooked visual clues and prevent planning within a
self-consistent but false model; conversely, exact frame matching may encourage
over-engineering or fitting incidental visual details. None of these trade-offs
is obvious a priori.

This article therefore studies a nested family of four agents that differ as
little as possible:

\begin{enumerate}[label=(\arabic*),leftmargin=2.2em]
  \item \twma{} maintains a textual world model, but is not required to
  maintain an executable simulator;
  \item \ewma{} adds the requirement to maintain an executable world model and
  planning code;
  \item \ewmas{} adds controller-triggered simplification and generalization
  passes; and
  \item \ewmasv{} adds a fixed executable-model interface, an exact
  observation-reproduction requirement, and verifier programs.
\end{enumerate}

These variants are configurations of the common system architecture shown in
\cref{fig:system-architecture} that differ in their prompts and initial
workspace contents. Their implementation-level differences are summarized in
\cref{tab:variants}.

Hereafter, we refer to these systems as the \emph{textual},
\emph{flexible-interface executable}, \emph{flexible-interface
simplification}, and \emph{fixed-interface verification} variants. For
brevity, we subsequently call them the textual, executable, simplification, and
verification variants, respectively.

This construction isolates three successive effects: adding an executable model
alongside the textual representation, adding explicit simplification to
executable modeling, and adding verification to simplified executable modeling.
The study follows a \emph{nested ablation ladder} and does not compare all
possible combinations of executable modeling, simplification, and verification.
In particular, we do not evaluate verification without simplification. Moreover,
the final comparison changes both the verification objective and the supporting
workspace: the verification variant receives fixed interfaces, templates, and
verifier programs. Its effect should therefore be
interpreted as the effect of the complete verification treatment rather than
as the isolated effect of a single instruction.

All four systems remain coding agents. In particular, the textual variant still
has shell and Python access and may run snippets in a REPL. In the textual
variant, only the requirement to maintain an executable world model as a
persistent deliverable is removed. This distinction is important: the
experiment asks whether a coding agent benefits from making its \emph{environment hypothesis} executable, not
whether the agent should be prohibited from using code at all.

We evaluate each agent with two language models, \gptfivefour{} and
\gptfivefive{}, and two reasoning-effort settings, \texttt{high} and
\texttt{xhigh}. The resulting $4\times2\times2$ design tests not only whether a
component helps on average, but also whether its value changes with model
capability or inference-time reasoning effort. Each of the 16
variant--model--effort combinations is evaluated once on each of the same 25
public games, with no cross-game memory.

The original \texttt{baseline1} agent was developed as a proof of concept and
accumulated several tightly coupled mechanisms, especially around verification.
Directly ablating that system would have produced variants with many incidental
differences. We therefore built the simpler v1.2 family used here. The design
goal of the verification variant is to retain performance in the range of the
original verification-equipped baseline while making the causal comparisons
easier to interpret; we do not claim that it is the strongest possible ARC-AGI-3 agent.

In simplifying the original \texttt{baseline1} agent into the v1.2 family, we
removed several ad hoc trouble-resolution mechanisms that were activated after
an unsuccessful level attempt. As a secondary, ceiling-oriented experiment, we
restore one such mechanism in a later version of the verification variant. This
follow-up changes more than one implementation detail and is therefore not
treated as a controlled estimate of the prompt's effect.

The verification variant's workspace still contains
\texttt{plan\_executor.py}, together with its other helper programs. Given a
proposed action sequence, this utility uses the world model to predict the next
observation and environment status, executes one action in the real environment,
compares the prediction with the actual result, and aborts the remaining
sequence on any mismatch. However, none of the agent-facing prompts mentions
the plan executor or requires its use.
It is therefore available for the agent to discover and use, but online plan
gating is not part of the instructed protocol. The main study focuses on
executable representation, simplification, and offline replay verification.

\section{Architecture and Agent Variants}
\label{sec:architecture}

\subsection{System Architecture}
\label{sec:system-architecture}

\begin{figure}[t]
  \centering
  \includegraphics[width=0.94\linewidth]{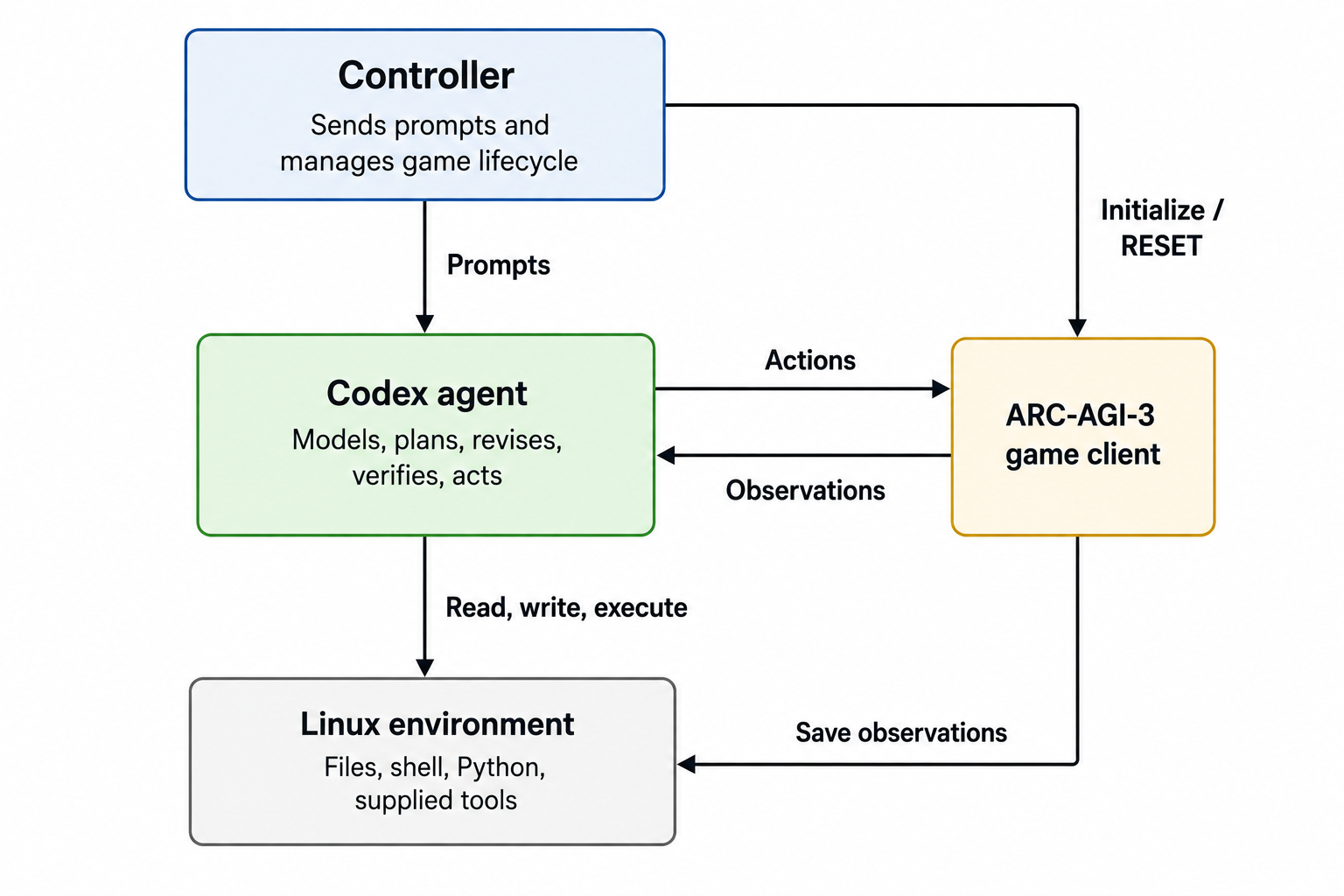}
  \caption{System architecture shared by all four variants. The controller
  sends prompts and manages the game lifecycle. A single Codex agent performs
  the substantive modeling, planning, artifact revision, program execution, and
  action selection. The ARC-AGI-3 client mediates interaction with the scored
  environment and saves observations in the Linux environment. The variants
  differ only in the prompts sent to Codex and the initial contents of its
  workspace; they do not change the architecture shown here.}
  \label{fig:system-architecture}
\end{figure}

All four evaluated variants use the same simple system architecture
(\cref{fig:system-architecture}). Its central component is a single Codex
coding agent operating in a Linux environment with access to persistent files,
a shell, and Python. A thin external controller sends prompts to Codex and
manages the game lifecycle. The ARC-AGI-3 client is a program included in the
agent's workspace; Codex runs it to submit actions and receive observations
from the current game. The workspace persists throughout a game playthrough,
but no workspace state is transferred between games.

The controller initializes the game, sends \texttt{RESET} through the client
when the current level must be restarted, and selects the next prompt sequence.
The controller itself contains no game-solving logic. All substantive
game-solving work is performed by Codex: it reasons about the game, works with
the files and programs in its workspace, and chooses all ordinary game actions.

Within this shared architecture, the variants differ in only two respects.
First, they receive different prompts: requirements for persistent artifacts
vary across variants, and the simplification and verification variants receive
additional simplification prompts. Second, only the verification variant
receives fixed-interface templates and verifier programs in its initial
workspace. All variants otherwise provide Codex with the same filesystem,
shell, and Python access and the same game-client program.

The executable environment model and planner are not separate agents or
independently running components. When required by the prompts, they are
programs developed and run by Codex in its workspace. Simplification is not an
autonomous module either; it is implemented through additional prompts sent to
the same agent. In the verification variant, verifier programs are included in
the initial workspace. Codex is instructed to run them, but the controller does
not enforce their use. None of these mechanisms adds another agent or
autonomous service.

\subsection{Controller Logic}
\label{sec:controller}

\begin{figure}[t]
  \centering
  \includegraphics[width=\linewidth]{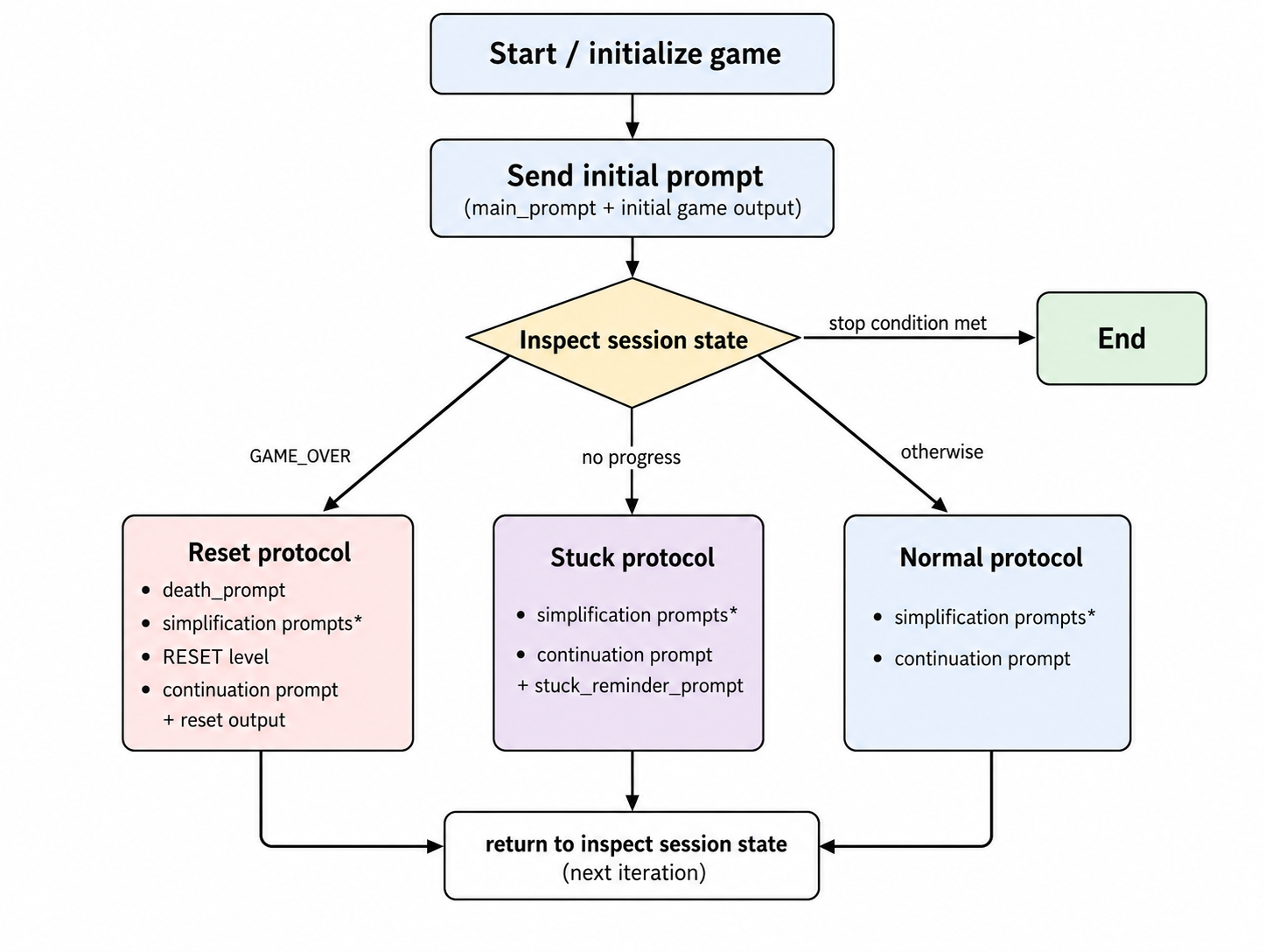}
  \caption{Controller state machine shared by the four agent variants. The
  simplification stages marked with an asterisk are enabled only for the
  simplification and verification variants. Stop conditions include solving the
  game, reaching the 1,500-action cap on the current level, having no active
  level, or making no
  progress after the stuck protocol. The runtime-recovery path is omitted for
  clarity.}
  \label{fig:controller}
\end{figure}

The controller follows the state machine shown in \cref{fig:controller}. It
bases its decisions only on basic game-progress information: the current game
status (\texttt{NOT\_FINISHED}, \texttt{GAME\_OVER}, or \texttt{WIN}); the
current level and attempt; the number of actions taken on the current level;
and whether Codex took any new actions during the preceding turn. This
information determines whether the controller stops or follows the reset,
normal, or stuck branch. The controller does not assess the content or quality
of the agent's hypothesis, executable model, planner, or verifier output.

All main-study playthroughs use Codex CLI version \texttt{0.128.0}
\cite{openai_codex_cli}. We treat the coding-agent runtime and its version as
part of the experimental system: changes to the runtime's built-in prompting,
context management, and tool-use behavior can change performance even when our
controller and prompts are held fixed.

At the start of a game, the controller obtains the initial observation from the
client. It then sends one initial message containing the
variant-specific main prompt, the common stop instruction, and the initial
client output. The same stop instruction is appended to every continuation
prompt: the agent is asked to work until it completes the current level or
reaches \texttt{GAME\_OVER}, and then return control to the controller. Within an
uninterrupted agent process, all later prompts resume the same Codex
conversation. The shared conversation is important because the
controller prompts are interventions in one continuing problem-solving process,
not independent calls to a stateless model.

When Codex returns control, the controller reads the recorded session state and
checks the stopping conditions. If the run should continue, it selects one of
three protocols shown in \cref{fig:controller}:

\paragraph{Normal protocol.}
If the game is not in \texttt{GAME\_OVER} and the agent has made at least one
new environment action since the preceding controller turn, the controller
optionally sends the simplification prompts and then sends the level-appropriate
continuation prompt.

\paragraph{Reset protocol.}
If the current attempt is in \texttt{GAME\_OVER}, the controller first sends a
death-analysis prompt asking the agent to explain the failure and update its
model. It then optionally runs the simplification stage, issues \texttt{RESET}
through the client, and sends a continuation prompt together with the reset
observation. The controller, rather than the coding agent, is intended to own
this reset transition.

\paragraph{Stuck protocol.}
If Codex returned without taking any new environment action, the controller
optionally runs the simplification stage and then sends the normal continuation
text plus an explicit reminder to make meaningful moves. If the agent returns a
second consecutive time with the same environment-step count, the run stops.
This branch is rare, but prevents an unproductive prompt loop from continuing
indefinitely.

\medskip
The requested boundaries are advisory rather than perfectly enforced by the
language model. Codex sometimes continues directly into the next level instead
of returning at level completion, and it can occasionally discover and invoke
\texttt{RESET} despite being instructed not to. Such deviations are handled as
part of normal operation.

\subsection{Simplification schedule}
\label{sec:simplification-schedule}

For the simplification and verification variants, the controller runs a
simplification step each time the agent returns control. On level 1, and during
the first attempt at level 2 while fewer than ten actions have been taken, the
controller sends one light simplification prompt. On later states it sends four
prompts sequentially: three increasingly broad passes over the model's ontology
and dynamics, followed by a planner-refactoring pass. The same schedule is used
in the normal, reset, stuck, and recovery paths. For the textual and executable
variants, the simplification stage is skipped in every path.

The prompts do not define a numerical complexity objective. They implement a
practical simplicity pressure suited to a coding agent: remove distinctions not
forced by evidence, replace case-by-case behavior with shared parameterized
rules, separate genuinely level-specific data from common mechanics, and keep
the textual and executable models aligned. The full prompt bodies are reproduced
in \cref{app:prompts}.

\subsection{The four variants}
\label{sec:variants}

\Cref{tab:variants} summarizes how the four experimental configurations differ
within the shared architecture shown in \cref{fig:system-architecture}. Here,
``variant'' denotes a configuration of the same system that differs in its
prompts and initial workspace contents, not a different arrangement of agents
or software components. The three variants without verification begin without
treatment-specific model or planner templates and without verifier programs.
The verification variant additionally receives fixed-interface templates and
supporting programs, but no game-specific rules, level layouts, or solutions.
All four variants otherwise provide Codex with the same filesystem, shell, and
Python access and the same game-client program.

\begin{table}[t]
  \centering
  \caption{Nested agent variants. All four use the same coding-agent runtime,
  game client, controller logic, action limits, and game-general prompts except
  for the differences shown.}
  \label{tab:variants}
  \small
  \begin{tabularx}{\linewidth}{@{}l>{\raggedright\arraybackslash}Xcc@{}}
    \toprule
    Variant & Required world-model representation & Simplification & Verification \\
    \midrule
    \twma & Textual model; no maintained executable simulator required & No & No \\
    \ewma & Text plus executable model and planning code; interfaces chosen by the agent & No & No \\
    \ewmas & Same as \ewma & Yes & No \\
    \ewmasv & Text plus executable engine, state reconstruction, renderer, and planner with fixed interfaces & Yes & Yes \\
    \bottomrule
  \end{tabularx}
\end{table}

\paragraph{Textual world-model variant (\twma).}
The textual variant is required to maintain a concise textual description of
its world model, but not executable world-model or planning code. Accordingly,
its main prompt contains no executable-model section or executable-code deliverable, and
its later-level continuation prompt does not ask the agent to update or create a
planner. The client, controller, and runtime are otherwise unchanged. The exact
main and continuation prompts are included in the appendix.

\paragraph{Flexible-interface executable world-model variant (\ewma).}
The executable variant differs from the textual variant only through a small
set of prompt changes that require the agent to maintain executable world-model
and planning code in addition to a textual description.
The agent may choose the state and action representations, file organization,
interfaces, and algorithms. It receives no fixed model templates, exact
observation-reproduction requirement, or verifier programs.

\paragraph{Flexible-interface executable world-model variant with
simplification (\ewmas).}
The simplification variant is identical to the executable variant except that
the controller runs the simplification procedure described in
\cref{sec:simplification-schedule}.

\paragraph{Fixed-interface executable world-model variant with simplification
and verification (\ewmasv).}
The verification variant starts from the simplification variant and replaces
the flexible-interface executable-model section with a fixed four-part contract: a
transition engine, initial-state reconstruction, an observation renderer, and a main planner. It also receives
workspace templates and general-purpose programs for replaying attempts,
reconstructing states, running planners, visualizing frames, and diagnosing
mismatches. Its main prompt instructs the agent to run the world-model and
planner verifiers after every modification to the executable model.

\subsection{Concrete example: a successful playthrough}
\label{sec:concrete-playthrough}
The common stop instruction, reproduced in \cref{app:stop-instruction}, is
included in the initial message and every continuation prompt
(\cref{app:composition}). The corresponding controller flow is shown in
\cref{fig:controller}. The instruction asks Codex to continue acting until it
completes the current level or reaches \texttt{GAME\_OVER}, and then return
control to the controller.

Consider an idealized playthrough in which Codex completes every level on its
first attempt and returns when requested. The controller sends the initial
message, Codex completes level 1, and control returns to the controller. In the
textual and executable variants, the controller then sends the appropriate
continuation prompt directly. In the simplification and verification variants,
it first sends the light simplification prompt and then sends the continuation
prompt. In all four variants, Codex is expected to complete level 2 before
returning control again.

After Codex completes level 2, the textual and executable variants again
proceed directly to their respective continuation prompts. In the
simplification and verification variants, the controller instead sends three
world-model simplification prompts followed by one planner-refactoring prompt
as four separate Codex turns, and then sends the continuation prompt. Each
simplification turn asks Codex to revise its model or planner;
it does not include an instruction to resume real-game actions. The intended
behavior is therefore for Codex to revise its workspace and return without
taking another game action, although this boundary is not mechanically
enforced. The same sequence repeats after later levels.

During game solving, Codex works differently according to the instructions for
each variant. In the textual variant, it solves the level while maintaining a
textual description of its current environment hypothesis in
\texttt{world\_model.md}. It can also use the shell and Python for local
analysis or situational planning, including running small transient snippets,
but it is not required to turn this code into a persistent simulator or
planner.

In the executable variant, Codex maintains both the textual description and
persistent executable environment-model and planning programs. It chooses their
organization and interfaces and is instructed to use them to support planning.
The controller does not prescribe how the model should represent a particular
game or inspect how the resulting programs contribute to each selected action.
The simplification variant follows the same game-solving procedure, but between
game-solving turns it additionally receives prompts asking it to simplify and
generalize its model and planner.

In the verification variant, Codex maintains the textual and executable models
and planner using the predefined interfaces detailed in
\cref{app:main-sv}. It is also instructed to use the planner to guide its
actions and to run the supplied verifier programs after modifying the
executable model. The prompt does not, however, prescribe a particular planning
algorithm; Codex chooses the algorithm and its implementation. Planner and
verifier use remains Codex's responsibility: the controller neither
invokes these programs automatically nor enforces their use before real actions
are taken.

In practical terms, a treatment is activated by changing the required
deliverables, inserting additional simplification prompts, or adding files to
the initial workspace. No treatment starts another agent or autonomous service.
The complete run artifacts, including the generated textual and executable
models, are available in the accompanying archive
\cite{rodionov2026supportingdata}.

\subsection{What the verifiers check}
\label{sec:verification}

The principal verifier replays every recorded attempt from level 1 through the
current level. For each attempt, it reconstructs the initial model state,
checks that the renderer reproduces the initial settled $64\times64$ ASCII
frame exactly, and then advances the model using the recorded action sequence.
At every step it compares the predicted environment status with the recorded
status. For nonterminal states it also requires a cell-exact match between the
rendered settled frame and the recorded settled frame. Intermediate animation
frames are available to the agent as evidence but are not part of this exact
replay test.

A second verifier checks the main planner on completed levels. It reconstructs
the initial state, asks the planner for an action sequence, simulates that
sequence in the executable model, and requires the simulated trajectory to end
in \texttt{LEVEL\_COMPLETED}. Additional programs allow the agent to run the
main or an auxiliary planner from the initial state, the current state, or an
intermediate point in a recorded attempt.

The renderer exposes a narrow override hook for unresolved visual details. Any
use of this hook is reported as a warning and is described in the prompt as
modeling debt rather than a satisfactory explanation. This escape hatch keeps a
partially understood model usable while preserving a visible signal that some
mechanic, latent variable, object identity, or rendering rule remains missing.

Verification is prompted and supported by supplied programs, but it is not
silently enforced by the external controller: the coding agent is responsible
for running the verifiers and responding to their diagnostics. Likewise, real
actions are not required to pass through an online plan executor in the main ablation.
Accordingly, the \ewmas{}--\ewmasv{} comparison evaluates the effect of the
complete verification treatment, not the isolated effect of a single
verification instruction.

\subsection{Runtime recovery}
\label{sec:recovery}

Coding-agent runs occasionally encounter technical failures. In our
experiments, Codex sometimes exited because of runtime or service errors; at
other times, its process remained alive but appeared to freeze and stopped
making observable progress. Because playthroughs may last many hours, treating
every such interruption as terminal would discard substantial progress. We
therefore use a supervisor that restarts the controller in recovery mode after
an unexpected process exit or after more than 30 minutes with no detected activity,
provided that no evaluation stopping condition has been reached.

Recovery restarts only the coding-agent process; the game server, recorded
environment trajectory, and workspace are preserved. Because the new Codex
process begins a new conversation rather than resuming the interrupted one, it
receives the main prompt together with a recovery prompt asking it to reconstruct
the current situation from the workspace and client state. It then enters the
normal continuation protocol, including the variant's simplification stage when
applicable. If the interruption occurred in \texttt{GAME\_OVER}, the controller
resets the current level attempt before continuing. At most ten recovery
attempts are allowed per game playthrough.

\section{Evaluation Harness and Protocol}
\label{sec:evaluation}

\subsection{Preventing unintended information access}
\label{sec:isolation}

A coding agent can inspect files and processes, execute arbitrary local code,
and search for unintended interfaces. The evaluation harness must therefore do
more than hide the game name in the natural-language prompt. As described in
our first article, early development harnesses exposed enough information for
an agent to recover the true game identifier, use web search, or start a second
local client as an unscored simulator. Earlier experiments revealed a clear
case of unintended information access and an apparent case of second-client
use. That article documents these incidents and the hardened harness introduced
to close the corresponding channels \cite{rodionov2026executableworldmodels}.
No results produced under
the vulnerable configurations are used in this study.

The current harness runs the agent in an isolated Docker environment and closes
the observed leakage channels. The true game identifier is absent from
agent-visible files, process arguments, environment variables, and API
responses. The agent-visible system also contains no reference to
\texttt{ARC} or \texttt{ARC-AGI}. General internet access and Codex web search
are disabled; a separate restricted proxy permits only the service endpoints
needed by the coding-agent runtime. Finally, the client--server protocol rejects
attempts to start a second game client. The same game-general client, prompts,
and controller are used for every game.

These measures are engineering controls, not a formal security proof against an
arbitrarily adversarial agent. One remaining attack vector would be for the
coding agent to modify its own configuration or source code in an attempt to
re-enable disabled capabilities such as web search. We have not observed such
behavior. The purpose of the controls is to close the concrete channels found
during the audit and to prevent the agent from identifying the benchmark or
public game by name.

These controls may also reduce the risk that the agent benefits from
benchmark-specific information memorized during training. This concern is
particularly relevant to \gptfivefive{} in the main study and
\gptfivesixsol{} in the follow-up, both of which were released after the public
ARC-AGI-3 games became available \cite{openai2026gpt55,openai2026gpt56}.
Because the agent is not told that the task is
related to ARC-AGI-3 and does not know the true game identifier, it receives
fewer cues with which to retrieve or recognize memorized benchmark-specific
information. This does not eliminate the possibility of contamination, since a
model could still recognize a game from its observations alone.

\subsection{Benchmark and evaluation design}
\label{sec:evaluation-design}

ARC-AGI-3 consists of interactive, level-based games in which the agent must
infer the mechanics and goals from observations and interaction. At each step,
the agent observes the current game state and submits one action from the
available action set. A level attempt ends either in level completion or in the
\texttt{GAME\_OVER} state. The benchmark-provided \texttt{RESET} action can be
used at any time to restart the current-level attempt and counts as an
environment action. \texttt{RESET} restarts only the current level; it cannot
return the agent to a completed level or restart the game from the beginning.
In our setup, Codex is instructed not to issue
\texttt{RESET} itself, although this instruction is not always followed;
resets are ordinarily issued by the controller after \texttt{GAME\_OVER} to
begin a new attempt at the current level.

The primary metric is Relative Human Action Efficiency (RHAE), the official
ARC-AGI-3 score \cite{arcprize2026arcagi3}. RHAE combines level completion with
action efficiency: for each completed level, the number of actions used is
compared with a human baseline, and the resulting level scores are aggregated
over the game and then across games.

We evaluate the four agent variants described in \cref{sec:variants} using
\gptfivefour{} and \gptfivefive{}, each with \texttt{high} and
\texttt{xhigh} reasoning effort. We use the term \emph{playthrough} for one
agent instance playing one game, and \emph{evaluation block} for the complete
sweep of 25 public games under one agent--model--effort condition.

Every playthrough starts from a fresh agent process and a clean workspace, with
no files, conversation state, or learned artifacts carried over from another
game. Each public game is played once per evaluation block.

As a resource-use proxy, we report \emph{cost tokens}. If
$T_{\mathrm{cached}}$, $T_{\mathrm{input}}$, and $T_{\mathrm{output}}$ are the
recorded cached-input, total-input, and output token counts, respectively, then
the cost-token count is
\[
 C = \frac{T_{\mathrm{cached}}}{60}
   + \frac{T_{\mathrm{input}}-T_{\mathrm{cached}}}{6}
   + T_{\mathrm{output}}.
\]
The weights normalize the base API token prices listed by OpenAI for both
models to the price of one output token \cite{openai2026apipricing}. Under those
prices, one million cost tokens would correspond to USD 15 for \gptfivefour{}
and USD 30 for \gptfivefive{}. We do not report converted dollar
estimates because the experiments used ChatGPT Pro-backed Codex accounts and
were not billed through the API. In preliminary API-key testing, prompt-cache
hit rates were substantially lower than in the subscription-backed runs,
shifting much of the input from cached to uncached tokens. Under the prices
above, uncached input costs ten times as much as cached input. Consequently,
the observed API cost was roughly five times our projection based on the cache
mix of the subscription-backed runs. Directly converting the token counts
reported here would therefore substantially underestimate the API cost of
reproducing these trajectories. Cost tokens should be
read as a consistent proxy for resources consumed by the recorded runs, not as
a billing forecast. For the GPT-5.6 follow-up, we retain the same formula only
to keep the resource proxy comparable; it is not a GPT-5.6 price model.

\subsection{Computational environment and runtime}
\label{sec:computational-environment}

All reported experiments were run on a Linux virtual machine with 32 vCPUs
(reported by the guest system as AMD EPYC 9454) and 64~GiB of RAM.
Playthroughs ran in isolated Docker containers using Docker version
\texttt{29.4.3}. To protect the host from rare runaway memory allocations in
programs generated by Codex, each agent process was launched inside its
container with \texttt{ulimit -v 6000000}. This sets a per-process
virtual-address-space limit of 6,000,000~KiB (about 5.7~GiB), inherited by
processes started by the agent. It is a safety cap, not a reservation or an
estimate of typical or required memory. We did not record peak per-container
memory, so the reported 64~GiB describes the available host memory rather than
a measured minimum requirement. The complete source code and Docker
configuration are openly available at
\url{https://github.com/astroseger/arc-3-agents-baseline1} and in the archived
release cited in the Code Availability statement.

The experimental campaign ran from 25 June to 15 July 2026, an elapsed interval
of approximately 19.5 days, with up to 18 playthroughs running concurrently.
Across the reported playthroughs, the median wall-clock runtime was 4~h 26~min,
the mean was 6~h 8~min, and the maximum was 53~h 8~min.

To complete the campaign within this interval, we used nine ChatGPT Pro
subscriptions, with a total subscription cost of USD~1,800. Based on the
observed usage, we estimate that four to five accounts would have been
sufficient to complete a 25-game evaluation block within their combined weekly
allowances, even for the most resource-intensive verification variant; the
textual variant required substantially less capacity. One account can run
several playthroughs per weekly allowance and can therefore execute a complete
block over multiple allowance periods. These estimates describe the service
limits observed during the study rather than guaranteed future capacity,
because subscription limits and model availability may change.

\subsection{Interaction rules and stopping conditions}
\label{sec:stopping}

A playthrough is one uninterrupted game trajectory across its sequence of
levels. The agent cannot restart the whole game to seek a better trajectory and
cannot return to a previously completed level.

The playthrough stops when the game is solved, when no active level remains,
when the agent makes no progress after the stuck reminder, or when the
evaluation-specific limit of 1,500 environment actions has been reached on the
current level. ARC-AGI-3 itself imposes no hard per-level action limit. Actions
taken after \texttt{GAME\_OVER} and before a reset are not free and remain part
of the recorded action budget.

The 1,500-action limit may be restrictive for some games. In the human-baseline
data for the 25 public games, the largest per-level baseline is 578 actions
(level 6 of \texttt{dc22}), and nine levels across six games have baselines of
at least 300 actions
\cite{arcprize2026humanbaseline,kamradt2026humanperformance}. Future evaluations
aimed at maximizing performance should use a higher per-level limit.

\subsection{All-or-nothing policy for irrecoverable failures}
\label{sec:failures}

We use an all-or-nothing policy at the evaluation-block level. If any of the 25
playthroughs in a block suffers an irrecoverable technical failure, none of the
playthroughs from that block enters the analysis; the complete 25-game block is
run again from scratch. Recoverable coding-agent interruptions are handled
within the original trajectory and do not trigger this policy.

This policy prevents selective reruns from biasing the reported performance.
Technical failure is not independent of performance: on the same game, a run in
which the agent struggles will generally take longer than one in which it
quickly discovers a solution. Longer trajectories have more opportunity to
encounter a service interruption, host failure, or usage limit. Replacing only
interrupted playthroughs could therefore preferentially discard difficult,
potentially low-scoring trajectories and bias the results upward.

The policy was triggered once. A host reboot interrupted the
\ewmasv{}--\gptfivefive{}--\texttt{xhigh} block before more than half of its
playthroughs had completed. We excluded every completed playthrough from that
partial block and restarted the entire 25-game block.

\section{Results}
\label{sec:results}

\subsection{Aggregate performance}
\label{sec:aggregate-results}

\begin{figure}[t]
  \centering
  \includegraphics[width=\linewidth]{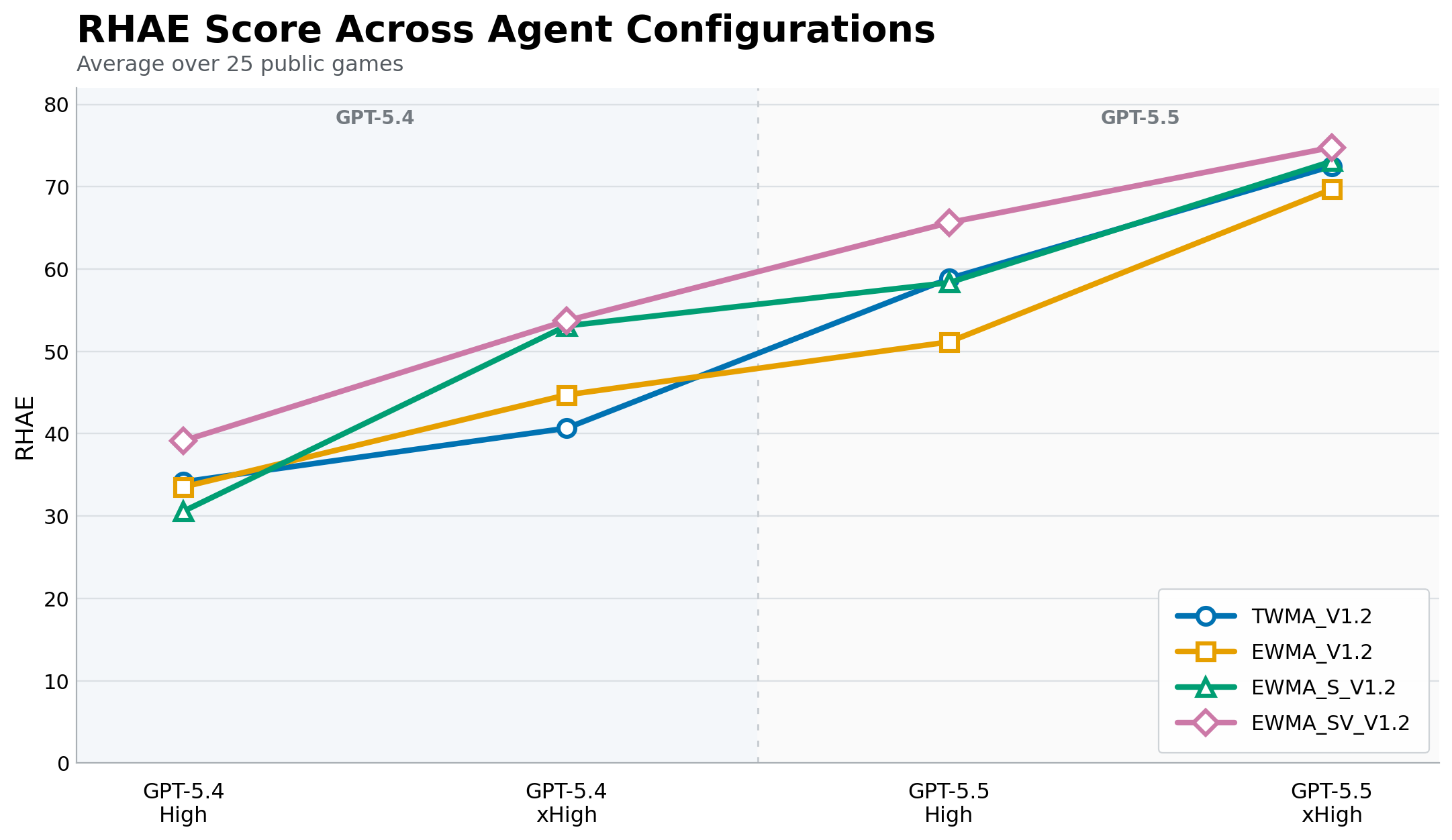}
  \caption{Mean RHAE across the 25 public ARC-AGI-3 games for the four main
  agent variants}
  \label{fig:main-results}
\end{figure}

\Cref{fig:main-results} plots the average RHAE over the 25 public games.
\Cref{tab:aggregate-results} reports the same RHAE values together with the
numbers of games and levels left unsolved. Full per-game results appear in
\cref{app:per-game-results}.

\begin{table}[t]
  \centering
  \caption{Aggregate main-study results. Each cell reports the RHAE score
  averaged over the 25 public games, followed in parentheses by the numbers of
  unsolved games and levels. Bold marks the highest RHAE within each
  model--effort setting.}
  \label{tab:aggregate-results}
  \small
  \begin{tabular}{@{}c@{}}
  \resizebox{\linewidth}{!}{%
  \begin{tabular}{@{}llrrrr@{}}
    \toprule
    Model & Effort & Textual & Executable & Simplification & Verification \\
    \midrule
    \gptfivefour & \texttt{high}  & 34.16 (19/85) & 33.54 (17/92) & 30.60 (20/88) & \textbf{39.16 (18/78)} \\
    \gptfivefour & \texttt{xhigh} & 40.67 (18/71) & 44.72 (16/73) & 53.10 (14/59) & \textbf{53.72 (13/45)} \\
    \gptfivefive & \texttt{high}  & 58.85 (13/36) & 51.16 (13/53) & 58.35 (11/34) & \textbf{65.64 (7/22)} \\
    \gptfivefive & \texttt{xhigh} & 72.51 (6/17)  & 69.70 (9/17)  & 73.09 (6/14)  & \textbf{74.78 (5/12)} \\
    \bottomrule
  \end{tabular}}
  \end{tabular}
\end{table}

The most uniform result is the effect of model capability and reasoning
effort. For every agent variant, \texttt{xhigh} outperforms \texttt{high}
within each model, and \gptfivefive{} outperforms \gptfivefour{} at each
matched effort. Averaging descriptively across the four variants gives RHAE
values of 34.36, 48.05, 58.50, and 72.52 for \gptfivefour{}--\texttt{high},
\gptfivefour{}--\texttt{xhigh}, \gptfivefive{}--\texttt{high}, and
\gptfivefive{}--\texttt{xhigh}, respectively. The number of levels left
unsolved declines along the same capability comparisons for every variant.

Differences among variants are less uniform. The verification variant has the
highest RHAE in all four model--effort settings, but its margin over the
simplification variant ranges from only 0.62 points with
\gptfivefour{}--\texttt{xhigh} to 8.56 points with
\gptfivefour{}--\texttt{high}. Simplification improves on the executable
variant in three settings, by 3.39--8.38 points, but is 2.94 points worse with
\gptfivefour{}--\texttt{high}. The executable requirement itself has mixed
results: relative to the textual variant, it changes RHAE by $-0.62$, $+4.05$,
$-7.69$, and $-2.81$ points across the four settings in table order.

\subsection{Resource use}
\label{sec:resource-results}

\Cref{tab:cost-results} reports the cost-token proxy defined in
\cref{sec:evaluation-design}. Simplification and verification are
resource-intensive: in every model--effort setting, the verification variant
uses the most cost tokens and the simplification variant uses the second most.
The textual and executable variants are substantially cheaper, although their
ordering depends on the model and effort.

\begin{table}[t]
  \centering
  \caption{Cost tokens, in millions, summed over the 25 public games. The
  values are a weighted token-use proxy defined in
  \cref{sec:evaluation-design}; they are not dollar costs.}
  \label{tab:cost-results}
  \small
  \begin{tabular}{@{}llrrrr@{}}
    \toprule
    Model & Effort & Textual & Executable & Simplification & Verification \\
    \midrule
    \gptfivefour & \texttt{high}  & 81.30 & 66.38  & 127.27 & 147.65 \\
    \gptfivefour & \texttt{xhigh} & 97.65 & 103.59 & 176.51 & 204.01 \\
    \gptfivefive & \texttt{high}  & 63.17 & 53.87  & 138.06 & 205.10 \\
    \gptfivefive & \texttt{xhigh} & 68.16 & 74.02  & 163.47 & 222.06 \\
    \bottomrule
  \end{tabular}
\end{table}

\section{Discussion}
\label{sec:discussion}

\subsection{Capability and reasoning effort dominate the main comparison}

The clearest conclusion is not about any one world-model representation. It is
that all four systems improve as the underlying model and inference-time
reasoning budget improve. Every one of the 16 matched capability comparisons
has the same direction: increasing reasoning effort helps each variant under
each model, and moving from \gptfivefour{} to \gptfivefive{} helps each variant
at each effort. No variant reaches a plateau within this controlled v1.2 range.

This supports the hypothesis that stronger coding models and larger reasoning
budgets would continue to improve performance across these architectures. That
claim is necessarily an extrapolation: two models and two effort settings
cannot establish the shape of a scaling curve or rule out a nearby ceiling.
What the experiment establishes directly is the unusually consistent direction
of the effect over the range tested.

\subsection{An executable deliverable is not universally beneficial}

At a fixed model and effort, all four variants remain in the same broad
performance regime. The within-setting range between the best and worst
variant is 5.08--14.48 RHAE points, whereas changing model or effort often
produces a larger shift. This was unexpected: a capable coding agent can remain
competitive even when instructed to maintain only a textual world model.

The result does not show that code is unnecessary. The textual variant is still
a coding agent with shell and Python access, and it can create ad hoc scripts.
Because the Codex conversation remains uninterrupted within a playthrough, it
can also retain and reuse code snippets from the conversation without turning
them into a maintained simulator. The ablation therefore concerns the value of
an executable world model as a persistent deliverable, not access to
computation.

Indeed, for \gptfivefive{} the textual variant outperforms the executable
variant at both efforts: 58.85 versus 51.16 at \texttt{high}, and 72.51 versus
69.70 at \texttt{xhigh}. It is instead comparable to the simplification
variant, which scores 58.35 and 73.09. In these conditions, requiring an
unconstrained executable simulator appears to divert effort from direct game
solving without providing a compensating advantage. With \gptfivefour{}, the
executable variant is essentially tied with the textual variant at
\texttt{high} and is 4.05 points better at \texttt{xhigh}. Thus, requiring an
executable world model as a persistent deliverable is not always beneficial,
at least under a flexible interface.

\subsection{Simplification usually helps; verification helps consistently}

The simplification variant scores higher than the executable-only variant in
three of four settings. The exception is the weakest tested coding-agent
condition,
\gptfivefour{}--\texttt{high}. This matches the failure mode we considered in
advance: repeated refactoring can be harmful when an agent has not yet formed a
stable model, especially when it must also invent the executable interface.
The result is nevertheless more favorable to simplification than this concern
suggested; at every stronger setting, explicit simplicity pressure helps.

We cannot infer whether simplification would help or harm the fixed-interface
verification variant under \gptfivefour{}--\texttt{high}, because the current
experiment does not include a verification-without-simplification variant.
Testing this combination is a natural direction for future work.

The verification variant ranks first in every setting. This is consistent with
the view that exact replay and explicit mismatch diagnostics improve problem
solving, but the advantage is smaller and less uniform than we initially
expected. At \gptfivefive{}--\texttt{xhigh}, for example, verification scores
74.78, while the textual and simplification variants already reach 72.51 and
73.09. One possible explanation is that agents without supplied verifiers
sometimes notice decisive visual cues directly or create their own checks when
uncertainty makes them useful. The present experiment does not test this
hypothesis. The cost-token results reinforce the practical trade-off: the
verification workflow gives the best aggregate score but is also the most
resource-intensive in every setting.

\subsection{Computational cost}
\label{sec:computational-cost}

The flexible-interface executable-model requirement alone does not impose a
consistent resource penalty: relative to the textual variant, it uses fewer
cost tokens in two model--effort settings and more in two. Scheduled
simplification and the complete verification treatment are consistently more
expensive. Across the four main-study settings, the verification variant uses
1.82--3.26 times as many cost tokens as the textual variant and 1.16--1.49
times as many as the simplification variant. Thus, its performance advantage
comes with substantially greater inference-time resource use.

Because playthroughs are mutually independent, they can be parallelized across
games, as illustrated by the 18-way concurrency reported in
\cref{sec:computational-environment}. This reduces the elapsed time of an
evaluation block but not its total model usage. We did not evaluate how
executable-model maintenance and replay verification scale to substantially
longer trajectories or more complex environments.

\subsection{Benchmark scope and external validity}
\label{sec:benchmark-limitations}

All evaluations in this study use the same 25 public ARC-AGI-3 games. This
allows controlled comparisons among matched agent configurations, but it is not
a held-out evaluation. The reported differences therefore apply to the
evaluated games and configurations; performance on ARC-AGI-3's semi-private and
fully private sets remains unknown.

Benchmark familiarity and training-data contamination cannot be ruled out. In
particular, \gptfivefive{} and \gptfivesixsol{} were released after the games
became public and may have encountered game-specific material during training.
The isolation controls in \cref{sec:isolation} omit the benchmark name and true
game identifier, reducing explicit retrieval cues. They cannot eliminate the
risk: a model may still recognize a game from its visual observations or
interaction dynamics and use previously encountered game-specific information.
The near-ceiling follow-up results should therefore be interpreted as
saturation of the public set only.

We did not evaluate substantially longer, noisier, less structured, or
real-world environments. The performance effects and resource costs of
executable world models, simplification, and verification may differ in such
settings; the present experiments do not establish how. Evaluating generality
requires held-out ARC-AGI-3 testing and corresponding ablations on other
interactive task families.

\subsection{Single-run variability and tunnel vision}
\label{sec:tunnel-vision}

Aggregate performance masks occasional severe failures. The most conspicuous
case is \texttt{r11l} under the verification variant with
\gptfivefive{}--\texttt{xhigh}: the playthrough scores only 4.76 and leaves five
levels unsolved. In the other three model--effort settings, the same variant scores
47.62, 100.00, and 85.93 on \texttt{r11l} and leaves at most two levels
unsolved; at \gptfivefive{}--\texttt{xhigh}, the textual and simplification
variants score 90.56 and 100.00. The result therefore appears to reflect a
transient, playthrough-specific tunnel-vision failure rather than a general
inability of the verification variant to solve the game.

The excluded block described in \cref{sec:failures} contained a second
qualitative warning. In that discarded block, the verification agent spent
1,008 actions on level 1 of \texttt{sk48}, whose human baseline is 61 actions
\cite{arcprize2026humanbaseline}, before eventually completing it. The block
was excluded under our all-or-nothing policy, but we mention this playthrough
because it helped
motivate the subsequent engineering intervention. Together, these observations
reinforce the run-to-run variability reported in our first article: performance
on an individual game can differ substantially across playthroughs
\cite{rodionov2026executableworldmodels}. Averaging over 25 games reduces the
influence of any one transient failure on the aggregate score, but a single
playthrough per game cannot estimate run-to-run variability for that game. These
observations also motivate the ceiling-oriented follow-ups below.

\section{Follow-up Experiments}
\label{sec:ceiling-experiments}

The main ablation was designed to assess the contribution of individual
components, not to maximize performance. We therefore ran five exploratory
evaluation blocks after the main
analysis: one v1.5 verification block and a matched $2\times2$ v1.6 comparison
of the textual and verification variants at two reasoning efforts. They use
the same 25 public games and reporting protocol. The v1.5 and v1.6
configurations differ from v1.2 in several components and are not additional
cells of the controlled main ablation. Within v1.6, however, the two variants
share those updates, so each effort provides a matched textual--verification
comparison. \Cref{tab:ceiling-results} summarizes the blocks;
per-game results appear in \cref{app:followup-per-game}.

\begin{table}[t]
  \centering
  \caption{Exploratory follow-up runs. UG/UL denotes unsolved games/unsolved
  levels, cost is millions of cost tokens, and actions is the sum of scorecard
  \texttt{level\_actions} values. Actions is reported only for blocks that
  completed all 183 levels; the corresponding sum of
  \texttt{level\_baseline\_actions} is 17,135. The v1.6 rows form matched
  textual--verification pairs within each effort; the v1.5 row is not part of
  those comparisons. Each row is one 25-game evaluation block.}
  \label{tab:ceiling-results}
  \small
  \begin{tabular}{@{}c@{}}
  \resizebox{\linewidth}{!}{%
  \begin{tabular}{@{}lllrrrr@{}}
    \toprule
    Variant & Model & Effort & RHAE & UG/UL & Actions & Cost \\
    \midrule
    \ewmasvonefive & \gptfivefive & \texttt{xhigh} & 82.01 & 5/9 & -- & 207.41 \\
    \addlinespace
    \twmavonesix   & \gptfivesixsol & \texttt{xhigh} & 92.34 & 2/5 & -- & 32.32 \\
    \ewmasvonesix & \gptfivesixsol & \texttt{xhigh} & 98.97 & 0/0 & 8,347 & 90.75 \\
    \twmavonesix   & \gptfivesixsol & \texttt{max}   & 95.97 & 0/0 & 10,111 & 30.60 \\
    \ewmasvonesix & \gptfivesixsol & \texttt{max} & 98.77 & 0/0 & 7,758 & 103.49 \\
    \bottomrule
  \end{tabular}}
  \end{tabular}
\end{table}

\subsection{Trouble-prompt add-back}
\label{sec:trouble-addback}

The \ewmasvonefive{} variant restores one of the anti-tunnel-vision mechanisms
from the original proof-of-concept agent. When the controller is handling a
reset after \texttt{GAME\_OVER}, it checks how many actions have been spent on
the current level. If more than 50 additional actions have accumulated since
the previous trouble intervention on that level, the controller sends an extra
prompt immediately before resetting. The level-1 prompt emphasizes that the
agent may be missing a simple visual cue; the later-level prompt asks what may
be missing from the world model. Both ask the agent to analyze prior attempts,
identify overlooked evidence, form a new hypothesis, and state a different
plan for the next attempt. The exact prompts are reproduced in
\cref{app:trouble-prompts}.

Version 1.5 also accelerated the game client provided to the agent. We did
not isolate this change experimentally. Consequently, in the
\gptfivefive{}--\texttt{xhigh} setting, the increase from 74.78 RHAE for the
main verification block to 82.01 for version 1.5 is a
ceiling-oriented systems result, not a controlled estimate of the effect of
either update. The v1.5 block leaves five games and nine levels unsolved.

\subsection{Matched GPT-5.6 Sol comparison}
\label{sec:gpt56-ceiling}

Version 1.6 applies the same harness updates to the textual and verification
variants: the v1.5 trouble intervention and accelerated game client, Codex CLI
\texttt{0.144.1} instead of \texttt{0.128.0}, and a client-side rule that
rejects all actions except \texttt{RESET} after \texttt{GAME\_OVER}. The CLI
update was required because version \texttt{0.128.0} did not support GPT-5.6
\cite{openai2026gpt56chatgpt}. We evaluate both variants with \gptfivesixsol{}
at \texttt{xhigh} and \texttt{max}
reasoning effort \cite{openai2026gpt56}.

At each effort, this provides a matched comparison of the textual and
verification variants under the same v1.6 harness. Comparisons with v1.5 or the
main study are not controlled across versions because both the model and
several harness components changed.

At \texttt{xhigh}, the verification variant scores 98.97 RHAE versus 92.34 for
the textual variant, a difference of 6.63 points. It completes all 183 public
levels, whereas the textual variant leaves five levels across two games
unsolved. At \texttt{max}, the corresponding scores are 98.77 and 95.97, a
2.80-point difference, and both variants complete all levels. The verification
variant also uses substantially more of the cost-token proxy: 90.75 versus
32.32 million at \texttt{xhigh}, and 103.49 versus 30.60 million at
\texttt{max}.

All three v1.6 blocks that complete all 183 levels use substantially fewer
actions than the summed human baseline of 17,135
\cite{arcprize2026humanbaseline}; their totals range from 7,758 to 10,111, a
reduction of 41--55\%. The two verification blocks use less than
half the human-baseline total despite scoring slightly below 100 RHAE.

The textual variant improves by 3.63 RHAE points from \texttt{xhigh} to
\texttt{max} and solves the five remaining levels, consistent with the
model-and-effort pattern in the main study. The verification variant is instead
near the measurement ceiling at both efforts. It solves every level in both blocks and
scores 100 on 23 of 25 games in each; the two sub-100 games are
\texttt{sc25} and \texttt{tn36} at \texttt{xhigh}, but \texttt{bp35} and
\texttt{s5i5} at \texttt{max}. In each case the residual loss is due to action
inefficiency on one or more completed levels relative to the human baselines,
not incomplete games. At \texttt{max}, the verification variant uses 7,758
actions, 589 fewer than at \texttt{xhigh}, even though its RHAE is 0.20 points
lower. Because RHAE depends on how actions are distributed across levels and
games rather than only on their grand total, these results are compatible. The
action totals therefore provide further evidence that the small reversal does
not show that \texttt{max} is worse; with one block per cell and scores pressed
against 100, the experiment cannot resolve such a small difference or estimate
run-to-run variability.

There is also a substantial contamination concern. GPT-5.6 was released after
the public ARC-AGI-3 games, so it may have encountered game-specific information
during training or evaluation. The isolation controls in
\cref{sec:isolation} remove the benchmark name and true game identifier, which
reduces direct retrieval cues, but a model could still recognize a game from
its visual observations. These results are qualitatively consistent with the
main study, but the near-ceiling scores on the public games do not establish
comparable performance on unseen games. Testing whether the performance
transfers requires a sealed or newly generated game set.

\section{Conclusion}
\label{sec:conclusion}

Contrary to the design intuition behind our original agent, our results suggest
that the answer to the question posed in the title is no---at least for
\gptfivesixsol{} at \texttt{max} on the public games. In that configuration,
the textual variant completed all 183 levels across the 25 public ARC-AGI-3
games while using 41\% fewer total actions than the summed human
first-contact baseline. It did so without being required to maintain a
persistent executable world model, receive scheduled simplification prompts,
or perform exact replay verification. The three imposed mechanisms were
therefore not necessary for action-efficient completion of this public set
under that configuration.

The results as a whole nevertheless give a more nuanced account of the
mechanisms' practical value. In the controlled main study, every variant scored
higher as model capability and reasoning effort increased, and these gains
often exceeded the within-setting differences among variants. Requiring a
persistent executable deliverable was not uniformly
beneficial: the textual variant scored higher than the flexible-interface
executable variant in both \gptfivefive{} settings. The simplification variant
scored higher than its executable-only counterpart in three of four settings,
with the weakest setting as the exception. The complete verification treatment
ranked first in all four settings, although its margin was sometimes small.

The matched \gptfivesixsol{} follow-up extends this pattern. The verification
variant completed every public level at \texttt{xhigh} and \texttt{max},
achieved approximately 99 RHAE, and used less than half as many total actions as
the summed human baseline. It scored higher than the textual variant at both
efforts and completed the set at \texttt{xhigh}, whereas the textual variant
completed it only at \texttt{max}. Across the main study and follow-up, the
verification variant ranked first in every matched comparison, but sometimes
only narrowly, and it used substantially more inference resources.

These comparisons have two important limitations. First, relative to the
simplification variant, the verification variant introduces a fixed
executable-model interface, supporting programs, and an exact replay requirement
in a single treatment. Its advantage therefore cannot be attributed to replay
verification alone.
Second, the study compares particular implementations rather than the maximum
attainable performance of each design family. In our exploratory follow-up at
the same model and effort, the improved verification system achieved a higher
score than the main-study version, directly demonstrating implementation
headroom for that variant. More broadly, this result suggests that other
variants may also have implementation headroom and that realizing it may
require improvements tailored to each design. In particular, targeted
improvements to the less resource-intensive textual variant might narrow or
even reverse its performance gap with the verification variant. We did not
test this possibility, so it remains a hypothesis for future work.

This conclusion should not be generalized beyond the evidence. All 25 games
evaluated here are public, and \gptfivesixsol{} was released after they became
available. Benchmark familiarity or training-data contamination therefore
cannot be ruled out. Results on these games are not a substitute for a sealed
held-out evaluation. Without such an evaluation, the study does not establish
the same answer for ARC-AGI-3's semi-private or fully private environments or
for substantially more complex real-world environments. In longer, noisier, or
less structured environments, the performance effects and resource overhead of
executable world models, simplification, and verification may differ from those
observed here. The present experiments do not establish the direction or
magnitude of those differences.

\paragraph{Future work.}
We identify three immediate directions.

First, the near-saturation observed on the public games should be tested on
held-out ARC-AGI-3 environments. A verified evaluation of the
\gptfivesixsol{} agents would test the hypothesis that Codex-based agents of
this strength are close to fully solving the ARC-AGI-3 benchmark, rather than
only saturating its public demonstration set.

Second, the two missing simplification comparisons should be run: a textual
variant with explicit simplification and a verification variant without it.
The first would test whether simplicity pressure helps a non-executable
representation. The second would isolate the contribution of simplification
within the verification treatment. Running this comparison across model
strengths and reasoning efforts would test whether fixed interfaces and replay
checks make simplification beneficial even under the weaker agent conditions
in which it did not help the flexible-interface executable variant.

Third, the same agent designs should be evaluated beyond ARC-AGI-3. The
underlying approach is not specific to this benchmark. AGI Maze is a
particularly relevant next testbed for partially observable, stateful tasks
\cite{potapov2026agimaze}. We can run evaluations on its held-out environments,
whereas we do not have access to the ARC-AGI-3 private sets. Repeating the same
ablations there would test whether the conclusions about simplification and
verification extend beyond the public ARC-AGI-3 games.

\backmatter

\section*{Declarations}

\bmhead{Funding}
This study received no specific grant. The work was conducted while the author
was employed as a researcher at SingularityNET.

\bmhead{Competing interests}
The author is employed by SingularityNET and declares no other financial or
non-financial competing interests.

\bmhead{Ethics approval and consent to participate}
Not applicable. This study involved no recruitment of or interaction with human
participants and used only publicly available aggregate human baseline
statistics; no identifiable human data were accessed.

\bmhead{Consent for publication}
Not applicable. This manuscript contains no identifiable personal data.

\bmhead{Data availability}
The complete result data and full run artifacts supporting this study are
available in Zenodo at
\url{https://doi.org/10.5281/zenodo.21412274}
\cite{rodionov2026supportingdata}. An additional index and mirror
links are available at
\url{https://github.com/astroseger/arc-3-agents-baseline1}.

\bmhead{Code availability}
The current agent code and prompts are available at
\url{https://github.com/astroseger/arc-3-agents-baseline1}. The version used
in this study is archived in Zenodo at
\url{https://doi.org/10.5281/zenodo.21417382}
\cite{rodionov2026arc3agentssoftware}.

\bmhead{Author contributions}
The author conceived and designed the study, developed the agent variants and
evaluation harness, conducted the experiments, analyzed the results, and wrote
and approved the manuscript.

\begin{appendices}
\setcounter{table}{4}
\renewcommand{\thetable}{\arabic{table}}
\section{Per-Game Results}
\label{app:per-game-results}

Tables~\ref{tab:per-game-54} and~\ref{tab:per-game-55} combine the two large per-game result matrices. Each cell
reports the final game score followed by the number of levels left unsolved
(\emph{score/UL}). Column prefixes T, E, S, and V denote the textual,
executable, simplification, and verification variants; H and X denote
\texttt{high} and \texttt{xhigh} reasoning effort.

\begin{table}[p]
  \centering
  \caption{Per-game results for \gptfivefour{}. Each cell is score/unsolved
  levels.}
  \label{tab:per-game-54}
  \scriptsize
  \setlength{\tabcolsep}{3.2pt}
  \begin{tabular}{@{}c@{}}
  \resizebox{\textwidth}{!}{%
  \begin{tabular}{@{}lrrrrrrrr@{}}
    \toprule
    Game & T-H & T-X & E-H & E-X & S-H & S-X & V-H & V-X \\
    \midrule
    \texttt{ar25} & 100.00/0 & 100.00/0 & 100.00/0 & 100.00/0 & 36.49/3 & 100.00/0 & 100.00/0 & 98.56/0 \\
    \texttt{bp35} & 0.07/8 & 1.25/8 & 0.31/8 & 9.03/6 & 0.36/8 & 1.57/8 & 0.43/8 & 24.67/3 \\
    \texttt{cd82} & 66.41/1 & 100.00/0 & 87.72/0 & 100.00/0 & 99.81/0 & 100.00/0 & 100.00/0 & 93.85/0 \\
    \texttt{cn04} & 4.76/5 & 77.72/0 & 4.76/5 & 0.00/6 & 47.24/2 & 81.21/0 & 46.94/1 & 9.82/4 \\
    \texttt{dc22} & 17.01/3 & 47.62/2 & 28.57/3 & 33.74/2 & 4.76/5 & 11.70/4 & 27.43/3 & 47.62/2 \\
    \texttt{ft09} & 27.18/0 & 35.89/2 & 89.57/0 & 100.00/0 & 5.22/2 & 57.99/0 & 14.29/4 & 80.21/0 \\
    \texttt{g50t} & 21.02/4 & 29.14/2 & 3.57/6 & 10.71/5 & 10.71/5 & 35.71/3 & 0.83/6 & 93.88/0 \\
    \texttt{ka59} & 30.29/1 & 25.17/1 & 30.24/1 & 28.72/1 & 25.92/2 & 30.33/1 & 16.59/3 & 40.21/2 \\
    \texttt{lf52} & 1.82/9 & 2.24/7 & 2.31/8 & 1.82/9 & 2.15/8 & 24.83/5 & 3.08/6 & 25.07/5 \\
    \texttt{lp85} & 100.00/0 & 100.00/0 & 85.12/0 & 100.00/0 & 80.63/0 & 100.00/0 & 100.00/0 & 77.10/0 \\
    \texttt{ls20} & 3.57/6 & 30.05/1 & 4.73/5 & 10.38/4 & 17.38/1 & 100.00/0 & 38.32/1 & 58.44/0 \\
    \texttt{m0r0} & 71.43/1 & 100.00/0 & 94.68/0 & 100.00/0 & 0.00/6 & 100.00/0 & 85.83/0 & 96.68/0 \\
    \texttt{r11l} & 47.62/2 & 17.16/3 & 0.69/5 & 47.62/2 & 47.16/2 & 47.62/2 & 47.62/2 & 100.00/0 \\
    \texttt{re86} & 46.97/2 & 51.06/1 & 48.35/2 & 27.78/4 & 48.20/2 & 52.75/2 & 36.16/3 & 48.57/2 \\
    \texttt{s5i5} & 8.16/4 & 4.22/6 & 0.46/6 & 11.28/4 & 16.67/5 & 4.76/5 & 30.86/2 & 58.33/2 \\
    \texttt{sb26} & 88.21/0 & 14.11/4 & 31.35/0 & 93.43/0 & 100.00/0 & 70.55/0 & 90.09/0 & 76.72/0 \\
    \texttt{sc25} & 0.03/4 & 0.00/6 & 0.00/6 & 0.00/6 & 0.00/6 & 21.08/3 & 5.49/4 & 0.00/6 \\
    \texttt{sk48} & 0.03/7 & 5.21/5 & 0.71/6 & 5.89/6 & 8.20/5 & 0.00/8 & 6.95/5 & 2.92/5 \\
    \texttt{sp80} & 4.76/5 & 4.76/5 & 4.76/5 & 21.04/3 & 28.57/3 & 47.62/2 & 4.76/5 & 11.83/4 \\
    \texttt{su15} & 2.22/8 & 2.97/6 & 2.22/8 & 2.96/7 & 6.81/6 & 2.22/8 & 1.87/8 & 21.53/5 \\
    \texttt{tn36} & 10.71/5 & 14.51/2 & 3.57/6 & 96.18/0 & 10.71/5 & 10.71/5 & 8.23/5 & 12.04/3 \\
    \texttt{tr87} & 88.79/0 & 100.00/0 & 100.00/0 & 84.63/0 & 47.30/0 & 100.00/0 & 100.00/0 & 100.00/0 \\
    \texttt{tu93} & 85.85/0 & 100.00/0 & 95.20/0 & 100.00/0 & 96.95/0 & 97.63/0 & 100.00/0 & 96.17/0 \\
    \texttt{vc33} & 23.64/3 & 53.57/2 & 18.44/4 & 16.80/3 & 21.43/4 & 100.00/0 & 12.40/4 & 34.82/0 \\
    \texttt{wa30} & 3.45/7 & 0.12/8 & 1.22/8 & 15.96/5 & 2.22/8 & 29.18/3 & 0.82/8 & 33.95/2 \\
    \bottomrule
  \end{tabular}}
  \end{tabular}
\end{table}

\begin{table}[p]
  \centering
  \caption{Per-game results for \gptfivefive{}. Each cell is score/unsolved
  levels.}
  \label{tab:per-game-55}
  \scriptsize
  \setlength{\tabcolsep}{3.2pt}
  \begin{tabular}{@{}c@{}}
  \resizebox{\textwidth}{!}{%
  \begin{tabular}{@{}lrrrrrrrr@{}}
    \toprule
    Game & T-H & T-X & E-H & E-X & S-H & S-X & V-H & V-X \\
    \midrule
    \texttt{ar25} & 100.00/0 & 100.00/0 & 99.43/0 & 100.00/0 & 95.60/0 & 100.00/0 & 100.00/0 & 100.00/0 \\
    \texttt{bp35} & 1.91/4 & 9.41/3 & 1.61/6 & 6.68/3 & 5.01/6 & 18.29/3 & 28.02/4 & 66.88/0 \\
    \texttt{cd82} & 95.22/0 & 100.00/0 & 82.03/0 & 93.59/0 & 100.00/0 & 94.84/0 & 100.00/0 & 100.00/0 \\
    \texttt{cn04} & 100.00/0 & 100.00/0 & 100.00/0 & 100.00/0 & 62.25/0 & 100.00/0 & 47.62/2 & 67.54/0 \\
    \texttt{dc22} & 47.62/2 & 100.00/0 & 45.29/2 & 47.62/2 & 42.57/2 & 47.62/2 & 49.21/1 & 55.30/1 \\
    \texttt{ft09} & 47.62/2 & 87.75/0 & 74.21/0 & 100.00/0 & 100.00/0 & 99.16/0 & 100.00/0 & 56.93/0 \\
    \texttt{g50t} & 22.84/2 & 39.11/0 & 8.49/5 & 66.58/0 & 35.71/3 & 33.05/0 & 66.78/0 & 85.86/0 \\
    \texttt{ka59} & 73.07/1 & 43.35/0 & 12.59/4 & 95.06/0 & 39.18/0 & 61.10/0 & 100.00/0 & 100.00/0 \\
    \texttt{lf52} & 13.80/6 & 30.45/4 & 9.41/6 & 16.79/4 & 17.32/5 & 72.13/0 & 25.81/4 & 75.13/0 \\
    \texttt{lp85} & 98.53/0 & 100.00/0 & 100.00/0 & 93.55/0 & 96.57/0 & 100.00/0 & 100.00/0 & 100.00/0 \\
    \texttt{ls20} & 26.83/3 & 43.63/0 & 20.92/0 & 54.24/0 & 9.08/2 & 87.24/0 & 66.47/0 & 65.61/0 \\
    \texttt{m0r0} & 100.00/0 & 100.00/0 & 100.00/0 & 100.00/0 & 100.00/0 & 100.00/0 & 100.00/0 & 100.00/0 \\
    \texttt{r11l} & 76.53/0 & 90.56/0 & 100.00/0 & 55.07/1 & 88.58/0 & 100.00/0 & 85.93/0 & 4.76/5 \\
    \texttt{re86} & 40.86/2 & 100.00/0 & 58.33/2 & 92.20/0 & 58.33/2 & 77.78/1 & 100.00/0 & 84.17/0 \\
    \texttt{s5i5} & 29.03/2 & 38.01/2 & 7.39/3 & 41.70/2 & 10.26/5 & 25.21/2 & 48.37/0 & 79.50/0 \\
    \texttt{sb26} & 100.00/0 & 100.00/0 & 100.00/0 & 88.84/0 & 44.12/0 & 97.19/0 & 76.76/0 & 88.50/0 \\
    \texttt{sc25} & 60.43/0 & 96.91/0 & 16.81/3 & 50.17/0 & 11.64/2 & 56.30/0 & 14.29/0 & 70.77/0 \\
    \texttt{sk48} & 16.67/5 & 16.67/5 & 16.67/5 & 53.65/1 & 73.31/0 & 20.60/4 & 35.79/0 & 65.17/0 \\
    \texttt{sp80} & 5.68/4 & 66.98/1 & 71.43/1 & 58.29/1 & 44.01/1 & 72.95/0 & 47.62/2 & 54.98/1 \\
    \texttt{su15} & 46.53/1 & 78.68/0 & 2.22/8 & 84.19/0 & 69.79/0 & 80.24/0 & 60.89/0 & 60.22/0 \\
    \texttt{tn36} & 83.96/0 & 73.80/0 & 79.09/0 & 65.10/0 & 84.19/0 & 63.83/0 & 8.89/5 & 75.00/1 \\
    \texttt{tr87} & 73.99/0 & 92.95/0 & 50.14/0 & 100.00/0 & 100.00/0 & 100.00/0 & 86.90/0 & 100.00/0 \\
    \texttt{tu93} & 96.01/0 & 100.00/0 & 93.77/0 & 96.17/0 & 93.86/0 & 92.74/0 & 100.00/0 & 100.00/0 \\
    \texttt{vc33} & 57.77/0 & 69.06/0 & 16.49/3 & 21.95/1 & 20.93/4 & 83.45/0 & 10.79/4 & 20.55/4 \\
    \texttt{wa30} & 56.38/2 & 35.37/2 & 12.56/5 & 61.03/2 & 56.40/2 & 43.59/2 & 80.89/0 & 92.70/0 \\
    \bottomrule
  \end{tabular}}
  \end{tabular}
\end{table}

\clearpage
\section{Follow-up Per-Game Results}
\label{app:followup-per-game}

Table~\ref{tab:followup-per-game} reports the exploratory per-game results.

\begin{table}[h]
  \centering
  \caption{Per-game results for the exploratory follow-up runs. Each result is
  score/unsolved levels. V1.5-X is the v1.5 verification variant with
  \gptfivefive{}--\texttt{xhigh}. T and V denote the v1.6 textual and
  verification variants, and X and M denote \gptfivesixsol{}--\texttt{xhigh} and
  \gptfivesixsol{}--\texttt{max}, respectively.}
  \label{tab:followup-per-game}
  \small
  \setlength{\tabcolsep}{4pt}
  \begin{tabular}{@{}lrrrrr@{}}
    \toprule
    Game & V1.5-X & T1.6-X & V1.6-X & T1.6-M & V1.6-M \\
    \midrule
    \texttt{ar25} & 100.00/0 & 100.00/0 & 100.00/0 & 100.00/0 & 100.00/0 \\
    \texttt{bp35} & 46.67/3 & 31.75/1 & 100.00/0 & 100.00/0 & 80.37/0 \\
    \texttt{cd82} & 100.00/0 & 100.00/0 & 100.00/0 & 100.00/0 & 100.00/0 \\
    \texttt{cn04} & 87.36/0 & 100.00/0 & 100.00/0 & 100.00/0 & 100.00/0 \\
    \texttt{dc22} & 71.43/1 & 100.00/0 & 100.00/0 & 96.47/0 & 100.00/0 \\
    \texttt{ft09} & 47.62/2 & 100.00/0 & 100.00/0 & 100.00/0 & 100.00/0 \\
    \texttt{g50t} & 85.86/0 & 100.00/0 & 100.00/0 & 92.29/0 & 100.00/0 \\
    \texttt{ka59} & 100.00/0 & 100.00/0 & 100.00/0 & 100.00/0 & 100.00/0 \\
    \texttt{lf52} & 90.36/0 & 32.80/4 & 100.00/0 & 100.00/0 & 100.00/0 \\
    \texttt{lp85} & 100.00/0 & 100.00/0 & 100.00/0 & 100.00/0 & 100.00/0 \\
    \texttt{ls20} & 98.62/0 & 100.00/0 & 100.00/0 & 82.15/0 & 100.00/0 \\
    \texttt{m0r0} & 100.00/0 & 100.00/0 & 100.00/0 & 100.00/0 & 100.00/0 \\
    \texttt{r11l} & 82.49/0 & 94.02/0 & 100.00/0 & 100.00/0 & 100.00/0 \\
    \texttt{re86} & 100.00/0 & 100.00/0 & 100.00/0 & 100.00/0 & 100.00/0 \\
    \texttt{s5i5} & 100.00/0 & 100.00/0 & 100.00/0 & 100.00/0 & 88.88/0 \\
    \texttt{sb26} & 100.00/0 & 100.00/0 & 100.00/0 & 100.00/0 & 100.00/0 \\
    \texttt{sc25} & 72.82/0 & 81.59/0 & 84.21/0 & 94.76/0 & 100.00/0 \\
    \texttt{sk48} & 81.94/0 & 86.61/0 & 100.00/0 & 100.00/0 & 100.00/0 \\
    \texttt{sp80} & 54.97/1 & 100.00/0 & 100.00/0 & 100.00/0 & 100.00/0 \\
    \texttt{su15} & 29.56/0 & 97.79/0 & 100.00/0 & 48.95/0 & 100.00/0 \\
    \texttt{tn36} & 90.58/0 & 100.00/0 & 90.10/0 & 94.97/0 & 100.00/0 \\
    \texttt{tr87} & 100.00/0 & 91.94/0 & 100.00/0 & 89.69/0 & 100.00/0 \\
    \texttt{tu93} & 99.26/0 & 97.74/0 & 100.00/0 & 100.00/0 & 100.00/0 \\
    \texttt{vc33} & 66.91/0 & 99.19/0 & 100.00/0 & 100.00/0 & 100.00/0 \\
    \texttt{wa30} & 43.73/2 & 95.16/0 & 100.00/0 & 100.00/0 & 100.00/0 \\
    \bottomrule
  \end{tabular}
\end{table}

\clearpage
\section{Exact Prompt Suite}
\label{app:prompts}

This appendix records the prompt bodies used by the v1.2 agents and, in its
final subsection, the two additional prompts used by the v1.5 and v1.6
follow-ups. Text is copied verbatim from the supplied agent directories,
including spelling and Markdown formatting. Typographic Unicode punctuation is
replaced with ASCII analogues for compatibility with the Springer Nature
template. Dynamic game observations are not reproduced; their insertion points
are specified below.

\subsection{Prompt composition}
\label{app:composition}

The initial controller message is
\begin{quote}
\nolinkurl{main_prompt} $+$ \nolinkurl{continuation_string} $+$
\texttt{"The initial output of the game client:"} $+$ the runtime client output.
\end{quote}
A normal continuation sends the level-specific continuation followed by
\nolinkurl{continuation_string}, preceded by the simplification sequence for the
simplification and verification variants. A reset transition sends
\nolinkurl{death_prompt}, then the optional simplification sequence, executes
\texttt{RESET}, and sends the normal continuation plus
\texttt{"The level has been reset. You have another attempt."}\allowbreak
\texttt{" Output from the client:"} and the new client output. A stuck transition
sends the optional simplification sequence and then the normal continuation plus
\nolinkurl{stuck_reminder_prompt}. Recovery starts a new conversation with
\nolinkurl{main_prompt} plus \nolinkurl{recovery_prompt}, after which the normal
protocol is applied.

The main-prompt mapping is shown in \cref{tab:prompt-mapping}.
\begin{table}[h]
\centering
\caption{Main-prompt and simplification-prompt mapping for the four agent variants.}
\label{tab:prompt-mapping}
\begin{tabular}{@{}lll@{}}
\toprule
Variant & Main prompt & Simplification prompts sent \\
\midrule
\twma & Appendix~\ref{app:main-twma} & No \\
\ewma & Appendix~\ref{app:main-ewma} & No \\
\ewmas & Appendix~\ref{app:main-ewma} & Yes \\
\ewmasv & Appendix~\ref{app:main-sv} & Yes \\
\bottomrule
\end{tabular}
\end{table}

\subsection{Main prompts}

\subsubsection{Main prompt for the textual variant (\twma)}
\label{app:main-twma}
\begin{Verbatim}
# Introduction

You are playing a game.

You interact with the real game through the client in `client`. Before doing anything else, read:

- `client/README.md`

Your primary objective is to build and maintain a world model of the game.

Use that world model to plan actions and complete all levels using as few moves as possible, including exploration moves.

Actions taken in the real game are costly because each move consumes a limited step budget. Simulation inside the world model is free.

The game consists of several levels with increasing complexity. Later levels usually extend earlier mechanics rather than replacing them, so a useful model may carry ideas forward across levels.

The client may return multiple frames for a single action because of animations or short transitions. For world-model work, final settled ASCII frames are usually the best frame source. Intermediate animation frames and PNG frames may still provide visual clues about mechanics.

# Attempts

If the game is in the `GAME_OVER` state, you should report it, and I will give you another attempt on the same level.

Remember that the game is deterministic: each attempt starts from exactly the same state, so the same actions will lead to exactly the same results.

Do not repeat the same mistakes. Think carefully, analyze previous failed attempts, and try a different approach in new attempts.

You are forbidden to call RESET by yourself.

# Required deliverables for each level

- `world_model.md` -- concise description of the current world model
- `level_N_reasoning_log.md` -- short log of hypotheses, evidence, observations, and model updates during level solving
- `level_N_report.md` -- short report written after solving the level, describing what you did and the final form of the model

# Helper programs

Use the following helper programs.

- `client` -- execute chosen in-attempt actions directly in the real game client.

# Notes

Keep `world_model.md` clear enough that you can understand what the model predicts, why you believe it, what objective you are pursuing, and what observations remain unresolved.

Keep `level_N_reasoning_log.md` brief but concrete.

After each level, write `level_N_report.md` with the main actions, useful clues, difficult observations, false hypotheses, and final model shape.

\end{Verbatim}

\subsubsection{Main prompt for the two flexible-interface variants
(\ewma{} and \ewmas)}
\label{app:main-ewma}
\begin{Verbatim}
# Introduction

You are playing a game.

You interact with the real game through the client in `client`. Before doing anything else, read:

- `client/README.md`

Your primary objective is to build and maintain an executable world model of the game.

Use that world model to plan actions and complete all levels using as few moves as possible, including exploration moves.

Actions taken in the real game are costly because each move consumes a limited step budget. Simulation inside the world model is free.

The game consists of several levels with increasing complexity. Later levels usually extend earlier mechanics rather than replacing them, so a useful model may carry ideas forward across levels.

The client may return multiple frames for a single action because of animations or short transitions. For world-model work, final settled ASCII frames are usually the best frame source. Intermediate animation frames and PNG frames may still provide visual clues about mechanics.

# Attempts

If the game is in the `GAME_OVER` state, you should report it, and I will give you another attempt on the same level.

Remember that the game is deterministic: each attempt starts from exactly the same state, so the same actions will lead to exactly the same results.

Do not repeat the same mistakes. Think carefully, analyze previous failed attempts, and try a different approach in new attempts.

You are forbidden to call RESET by yourself.

# Required deliverables for each level

- `world_model.md` -- concise description of the current world model
- executable world-model code, including planning code. You may choose the file organization and interfaces.
- `level_N_reasoning_log.md` -- short log of hypotheses, evidence, observations, and model updates during level solving
- `level_N_report.md` -- short report written after solving the level, describing what you did and the final form of the model

# Executable world model

The executable world model should be explicit enough to support planning. You may choose its internal state representation, action representation, algorithms, helper artifacts, and planner design.

The model should help you reason about candidate action sequences, terminal outcomes, and useful plans.

Keep the implementation as simple as you can while preserving useful planning.

# Helper programs

Use the following helper programs.

- `client` -- execute chosen in-attempt actions directly in the real game client.

## Useful Python libraries

Incorporate Python libraries like NumPy, SciPy, Sympy and NetworkX when they help your world model or planners.

# Notes

Keep `world_model.md` clear enough that you can understand what the model predicts, why you believe it, what objective you are pursuing, and what observations remain unresolved.

Keep `level_N_reasoning_log.md` brief but concrete.

After each level, write `level_N_report.md` with the main actions, useful clues, difficult observations, false hypotheses, and final model shape.

\end{Verbatim}

\subsubsection{Main prompt for the fixed-interface verification variant
(\ewmasv)}
\label{app:main-sv}
\begin{Verbatim}
# Introduction

You are playing a game.

You interact with the real game through the client in `client`. Before doing anything else, read:

- `client/README.md`

Your primary objective is to build and maintain an executable world model of the game.

Use that world model to plan actions and complete all levels using as few moves as possible, including exploration moves.

Actions taken in the real game are costly because each move consumes a limited step budget. Simulation inside the world model is free.

The game consists of several levels with increasing complexity. Later levels usually extend earlier mechanics rather than replacing them, so a useful model may carry ideas forward across levels.

The client may return multiple frames for a single action because of animations or short transitions. For world-model work, final settled ASCII frames are usually the best frame source. Intermediate animation frames and PNG frames may still provide visual clues about mechanics.

# Attempts

If the game is in the `GAME_OVER` state, you should report it, and I will give you another attempt on the same level.

Remember that the game is deterministic: each attempt starts from exactly the same state, so the same actions will lead to exactly the same results.

Do not repeat the same mistakes. Think carefully, analyze previous failed attempts, and try a different approach in new attempts.

You are forbidden to call RESET by yourself.

# Required deliverables for each level

- `world_model.md` -- concise description of the current world model
- `world_model_engine.py` -- world model engine, updated continuously so it remains valid for all solved levels so far
- `world_model_state_io.py` -- initial-state reconstruction and observation rendering, updated continuously so it remains valid for all solved levels so far
- `world_model_main_planner.py` -- main planner, updated continuously so it remains valid for all solved levels so far
- optional `level_N_planner_i.py` -- additional intermediate-state planners used during solving
- `level_N_reasoning_log.md` -- short log of hypotheses, evidence, observations, and model updates during level solving
- `level_N_report.md` -- short report written after solving the level, describing what you did and the final form of the model


# Executable world model

Your executable world model should have four explicit parts.

## 1. World model engine

This is the internal game dynamics model.

It should:

- represent the internal game state
- receive an action
- update the state according to inferred mechanics
- be as simple and general as possible

The world model engine must be implemented in `world_model_engine.py` as a function called `world_model_engine`.

The world model engine should model only the dynamics within a single attempt. It should not model transitions between attempts.

`world_model_engine` must receive exactly two arguments:

- `state` -- the current internal world-model state
- `action` -- the action to apply

and it must return:

- `new_state` -- the new internal world-model state
- `game_status` -- one of `RUNNING`, `LEVEL_COMPLETED`, or `GAME_OVER` (`RUNNING` is non-terminal state, `LEVEL_COMPLETED` and `GAME_OVER` are terminal states for an-attempt)

The engine state must be a dictionary with the following obligatory field:

- `level` -- the level index

and it must also contain the internal representation of the non-terminal game state.

The `action` must be represented as a dictionary with field `name` and two optional parameters `x` and `y` (for `ACTION6`).

Your goal is to infer a compact underlying rule system. The real mechanics are usually relatively simple.

Do **not** hardcode level layouts or ad hoc special cases into the engine unless absolutely necessary.

If some information really is level-specific, isolate it in a clearly separated level-specific data structure rather than burying it in the engine logic.

Hardcode as little as possible.
The model must explain observations through general mechanics, not by memorizing level-specific behavior.

It is strictly forbidden to load real game observations into the world model engine in any way.

## 2. State reconstruction

This logic reconstructs the initial state of the internal world model for each level.

It must be implemented in `world_model_state_io.py` with the following functions.

### `initial_state_reconstruction`

This function reconstructs the initial state for a given level.

The game is deterministic at level start: all attempts for the same level begin from the same initial world state and produce the same initial observation. This is true even in partially visible worlds.

It must receive exactly two parameters:

- `level_index` -- the index of the level
- `initial_frame` -- the initial settled ASCII frame of the current level

For some games, the entire world is visible in the initial frame; we call these "fully visible worlds." In such cases, you should reconstruct the world state from the provided initial frame.

For partially visible worlds, `initial_frame` may not contain enough information to reconstruct the full initial world state.

In such cases, you may choose any clean reconstruction strategy, but keep it separate from `initial_state_reconstruction(...)`. One acceptable approach is to write separate reconstruction code that uses all available observations from all attempts of the same level to estimate the level-wide initial full frame. For sliding or scrolling worlds, this full frame may be larger than the visible `64x64` frame.

If you use this approach, save the reconstructed full initial frame as a level artifact, and let `initial_state_reconstruction(...)` load only that artifact, not raw attempts, session logs, later observations, or other real game data.

The saved full initial frame is still a hypothesis. Be conservative at visibility boundaries: do not treat unseen regions as known unless they are forced by evidence.

It is strictly forbidden to load real game observations into `initial_state_reconstruction` in any way, except for the explicitly provided initial frame and any explicitly saved level-wide initial-state reconstruction artifact.

### State Reconstruction Principles

To reconstruct any later non-terminal state of the same attempt:

1. reconstruct the initial state with `initial_state_reconstruction(...)`
2. advance it with `world_model_engine(...)` by simulating the known attempt actions up to the required prefix

## 3. Observation renderer

This is the function that translates the internal world-model state back into the expected ASCII frame.

It must be implemented in `world_model_state_io.py` as a function called `state_renderer`.
You may also define a separate helper such as `apply_render_overrides(frame, state, level_index, attempt_index, step_count)` for verification-only temporary frame patches.

`state_renderer` must receive exactly one argument:

- `state` -- the internal world-model state produced by the world model engine

and it must return the corresponding ASCII frame as `numpy.ndarray` of shape `64x64`, dtype `np.int16`.

`state_renderer` only needs to render non-terminal level states. You do not need to render `LEVEL_COMPLETED` or `GAME_OVER` states.

This renderer is required so the model can be verified directly against real observations.

Your model is only acceptable if its rendered settled frame matches the real settled ASCII frame exactly.

If you absolutely cannot yet explain some frame-local visual detail, you may use `apply_render_overrides(...)` as a last-resort verification-only patch hook in `world_model_state_io.py`.
Use it only for narrow visual corrections to specific unresolved frames. Do **not** put game logic, state transitions, or planning logic into this hook.
Every `apply_render_overrides(...)` patch must be treated as temporary and as evidence that the world model is still missing a real mechanic, object identity, latent state variable, or observation rule.

It is strictly forbidden to load real game observations into the observation renderer in any way.

## 4. Main planner

The world model must support planning. The planner is the planner in terms of your internal world model.

It must be implemented in `world_model_main_planner.py` as a function called `planner`.

`planner` must receive:

- `state` -- the internal world-model state

and it must return either:

- a list of actions to reach level completion
- `None` if level completion is not reachable

It should be possible to use the world model engine, together with this planner, to infer a sequence of actions required to reach a desired game state.

In most cases, a simple search or planning algorithm should be sufficient.

You must use a planner to guide your actions.

- You must implement the main planner as soon as possible, usually by extending the current planner from the previous solved level.
- If your world model is already good enough, you must try to use `planner` to plan all the way to level completion. You may use `python3 run_main_planner.py --from-current` to try it from the current in-attempt state.
- If level completion is not yet reachable, prefer using a separate planner for intermediate exploration targets. If you do so, you must save it as `level_N_planner_i.py` for future inspection.
- Auxiliary planners may optionally accept `goal`: `planner(state, goal=None)`, where `goal` is a JSON-serializable dictionary describing the requested target.
- The auxiliary planner may share logic with `world_model_main_planner.py`, but it must still plan in terms of the explicit world model.
- After level completion, `world_model_main_planner.py` is a required deliverable and you must verify it with `python3 verify_main_planner.py`. You may also inspect the planner output with `python3 run_main_planner.py --from-initial N`.

Pathfinding or planning done outside the explicit world model is not acceptable.

# Helper programs

Use the following helper programs.

- `client` -- execute chosen in-attempt actions directly in the real game client.
- `python3 verify_world_model.py` -- verify the world model for levels `1..current_level` against all recorded attempts. For every attempt, it reconstructs the initial state, simulates the full attempt from that state, and checks the predicted status and rendered settled frame at each simulated step. If `apply_render_overrides(...)` changes a rendered frame, the verifier prints a warning; treat that as unresolved modeling debt and as a likely clue to the puzzle.
- `python3 verify_main_planner.py` -- verify the main planner for all completed levels. For each completed level, it reconstructs the initial state, runs the planner, and checks with the world model engine that the resulting plan reaches `LEVEL_COMPLETED`. It does not verify the current level unless that level has already been completed.
- `python3 run_main_planner.py --from-current`, `python3 run_main_planner.py --from-initial N`, or `python3 run_main_planner.py --from-attempt N A S` -- run the main planner and print the resulting action sequence. `--from-current` plans from the latest known state of the current level. `--from-initial N` plans from the initial state of level `N`. `--from-attempt N A S` plans from the state obtained by reconstructing the initial state and simulating the first `S` steps of attempt `A` on level `N`.
- `python3 run_aux_planner.py planner_module_name --from-current`, `python3 run_aux_planner.py planner_module_name --from-initial N`, or `python3 run_aux_planner.py planner_module_name --from-attempt N A S` -- run an auxiliary planner module such as `level_N_planner_i` and print the resulting action sequence. Auxiliary planners must expose a function called `planner`. They may also receive `--goal 'JSON'`, which is parsed and passed to the auxiliary planner as `goal`. If you need an exploratory or intermediate target, use a separate planner saved as `level_N_planner_i.py`, and run it with `python3 run_aux_planner.py ...`.

## Useful Python helpers

You may also inspect data or test ideas directly in a Python REPL.

- `from state_reconstruction_tools import reconstruct_initial_state` -- reconstruct the initial world-model state for a given level.
- `from state_reconstruction_tools import reconstruct_current_state` -- reconstruct the current world-model state from the latest recorded attempt of the current level.
- `from state_reconstruction_tools import reconstruct_state` -- reconstruct the world-model state for `level`, `attempt_index`, `step_count`.
- `from state_reconstruction_tools import simulate_actions` -- simulate a sequence of actions in the world model from a given state.
- `from frame_plot_lib import save_ascii_frame_png` -- render any 2D ASCII-frame `numpy.ndarray` to PNG. It also works for arbitrary subregions or stitched multi-frame panoramas, not just full `64x64` frames.
- `from session_tools import read_all_attempts_for_level` -- read all recorded attempts for a level.
- `from session_tools import read_attempt_for_level` -- read a specific recorded attempt for a level.
- `from session_tools import read_current_attempt` -- read the latest recorded attempt of the current level.

## Useful Python libraries

Incorporate Python libraries like NumPy, SciPy, Sympy and NetworkX when they help your world model or planners.

# Verification

After each modification to the executable world model, you should run both verifiers:

- `python3 verify_world_model.py`
- `python3 verify_main_planner.py`

# Notes

Keep `world_model.md` clear enough that you can understand what the model predicts, why you believe it, what objective you are pursuing, and what observations remain unresolved.

Keep `level_N_reasoning_log.md` brief but concrete.

After each level, write `level_N_report.md` with the main actions, useful clues, difficult observations, false hypotheses, and final model shape.

\end{Verbatim}

\subsection{Continuation, failure, stuck, and recovery prompts}

\subsubsection{Common stop instruction
(\texttt{continuation\_string.txt})}
\label{app:stop-instruction}
\begin{Verbatim}
# When to stop

Continue working until you either complete the current level or the game reaches the GAME_OVER state. When either condition occurs, stop taking actions in the real game, prepare the required deliverables, and report back to me.
\end{Verbatim}

\promptfile{Level-1 continuation (all variants)}{continuation_level1.txt}

\promptfile{Later-level continuation for the executable-model variants}{continuation_l2.txt}

\promptfile{Later-level continuation for the textual variant}{continuation_l2_twma.txt}

\promptfile{Death-analysis prompt}{death_prompt.txt}

\promptfile{Stuck reminder}{stuck_reminder_prompt.txt}

\promptfile{Recovery prompt}{recovery_prompt.txt}

\subsection{Simplification prompts}
\label{app:simplification-prompts}

These prompts are sent only for the simplification and verification variants.
The light prompt is used on level 1 and early in the first attempt at level 2.
In later states, the controller sends the three simplification prompts and then the planner prompt
as four separate, sequential messages.

\promptfile{Light simplification prompt}{light_simplification_level1.txt}

\promptfile{Simplification step 1}{world_model_simplification_step1.txt}

\promptfile{Simplification step 2}{world_model_simplification_step2.txt}

\promptfile{Simplification step 3}{world_model_simplification_step3.txt}

\promptfile{Planner-refactoring prompt}{world_model_planner.txt}

\subsection{Follow-up trouble prompts}
\label{app:trouble-prompts}

These two prompts are not part of the main v1.2 comparison. The v1.5 and v1.6
controllers use the level-1 version on level 1 and the general version on later
levels under the trigger described in \cref{sec:trouble-addback}.

\promptfile{Level-1 trouble prompt}{trouble1_prompt_level1.txt}

\promptfile{Later-level trouble prompt}{trouble1_prompt.txt}

\end{appendices}

\bibliography{arc3_article2_references}

\end{document}